\documentclass[lettersize,journal]{IEEEtran}
\usepackage{amsmath,amsfonts}
\usepackage{algorithmic}
\usepackage{algorithm}
\usepackage{array}
\usepackage[caption=false,font=normalsize,labelfont=sf,textfont=sf]{subfig}
\usepackage{textcomp}
\usepackage{stfloats}
\usepackage{url}
\usepackage{verbatim}
\usepackage{graphicx}
\usepackage{cite}
\usepackage{multirow}
\usepackage{stfloats} 
\usepackage{xcolor}
\usepackage{booktabs}
\usepackage{hyperref}
\usepackage{makecell}
\usepackage{amsmath}
\newtheorem{theorem}{Theorem}
\newtheorem{lemma}[theorem]{Lemma}

\usepackage{listings}
\definecolor{codegreen}{rgb}{0,0.5,0}
\definecolor{codeblue}{rgb}{0.25,0.5,0.5}
\definecolor{codegray}{rgb}{0.6,0.6,0.6}

\begin{document}

\title{Iwin Transformer: Hierarchical Vision Transformer \\ 
using Interleaved Windows}

\author{Simin Huo, Ning Li
        % <-this % stops a space
}



\maketitle

\begin{abstract}
Vision Transformers (ViTs) face two limitations: the rigid resolution dependency of positional embeddings, which complicates cross‑resolution fine‑tuning, and the quadratic complexity of attention. While Swin Transformer alleviates the latter through window attention, it suffers from fine-tuning. Following the philosophy "no token is an island," we present Iwin Transformer, a position‑embedding‑free hierarchical vision transformer that couples interleaved window attention with depthwise convolution inside a single block. Attention captures long‑range dependencies, while convolution links local neighbors and implicitly encodes spatial position. This design not only reduces the quadratic complexity of attention but also enables two types of scalability: fine‑tuning from low to high resolution and weight transfer from 2D to 3D. With window‑size adjustment alone, direct $224^2{\rightarrow}384^2$ fine‑tuning lifts Iwin‑L from 86.4\% to 87.4\% top‑1 accuracy on ImageNet‑1K. Transferring an ImageNet‑pretrained Iwin‑T to video achieves 79.1\% on Kinetics‑400, outperforming Swin‑T (78.8\%) with 15.9\% fewer FLOPs. Iwin also remains competitive on ADE20K segmentation and class‑conditional image generation (FlashDiT). Overall, Iwin offers an effective approach to simultaneously tackling the complexity and scalability challenges in ViTs. Code and models are at \href{https://github.com/cominder/Iwin-Transformer}{https://github.com/cominder/Iwin-Transformer}.
\end{abstract}


\section{Introduction}
IEEEPARstart{V}{ision} Transformers (ViTs)~\cite{dosovitskiy2020image} have transformed computer vision by applying self-attention~\cite{vaswani2017attention} to image patches. Unlike CNNs~\cite{krizhevsky2012imagenet}, ViTs model global dependencies but incur $\mathcal{O}(N^2)$ cost in sequence length $N$, limiting high-resolution applications. Numerous strategies, hierarchical pyramids, sparse attention, and CNN-Transformer hybrids, have been proposed to improve efficiency; see Sec.~\ref{sec:related}. Among window-based methods, Swin Transformer~\cite{liu2021swin} restricts attention to local windows and connects windows via shifting, achieving linear complexity while preserving hierarchical multi-scale features for dense prediction.

Despite its pioneering design and impressive performance, the Swin Transformer exhibits three noteworthy limitations. First, the shifted window mechanism introduces extra computational overhead due to the complex masking operations required during attention computation, complicating implementation and reducing hardware efficiency. Second, Swin's architecture requires two consecutive transformer blocks: one with regular windows and another with shifted windows to achieve global information exchange, resulting in computational redundancy as certain features are processed multiple times. This two-block requirement poses particular challenges in the era of AI generated content (AIGC), where conditioning information such as text prompts must be injected into the model; there is no obvious optimal placement for cross-attention between text and images within this rigid two-block structure, explaining Swin's limited adoption in modern text-to-image diffusion models. Furthermore, as acknowledged in Swin Transformer v2~\cite{liu2022swin}, the model faces scalability issues when fine-tuned for higher-resolution inputs. The bi-cubic interpolation of relative position encodings for larger windows leads to significant performance degradation, necessitating the introduction of complex alternatives such as log-spaced continuous position bias (Log-CPB). This reliance on sophisticated position encoding schemes ultimately hinders the model's scaling capabilities and broader applicability.

To address these limitations while preserving the efficiency of window-based attention, we introduce the \textbf{I}nterleaved \textbf{Win}dow Transformer (Iwin Transformer). The core innovation lies in coupling depthwise convolution with interleaved window attention, which rearranges tokens into an interleaved order before window attention such that each window contains tokens from diverse spatial regions of the image. This elegant design reduces quadratic complexity while achieving global information flow: any two tokens can interact either directly within a shared window or convolution kernel, or indirectly through an intermediate token acting as a bridge. Beyond global connectivity, this design yields two important byproducts. First, convolution introduces vision-friendly inductive biases that enrich local feature representation. Second, it supplies implicit positional information, eliminating the reliance on explicit positional encodings and thereby strengthening Iwin's scalability, directly addressing a key limitation of the Swin Transformer and making Iwin particularly well-suited for high-resolution and cross-resolution vision applications. Furthermore, the hybrid attention-convolution block functions as a self-contained, modular unit that can be seamlessly incorporated into multi-modal and generative model architectures.

Our main contributions are as follows:

\begin{enumerate}

\item \textbf{Interleaved window attention (IWA)}: We propose a novel Reshape-Transpose-Reshape (RTR) operation that reorders feature sequences into an interleaved pattern, then applies window attention, and finally restores the original spatial arrangement.

\item \textbf{Hybrid module and design guideline}: We integrate depthwise convolution with IWA to build global connectivity and prove a sufficient condition $KM \geq \max(H, W)$ that guides kernel/window selection.

% \item \textbf{Theoretical analysis}: We provide mathematical proof that Iwin achieves global information exchange through hybrid attention-convolution module.

\item \textbf{Position-embedding-free scalability}: Without position encodings, Iwin supports cross-resolution fine-tuning and weight transfer, significantly enhancing scalability. 

% \item  \textbf{Comprehensive empirical validation}: We provide extensive experimental evidence showing that Iwin maintains or improves upon the performance of Swin across various vision tasks including image classification, semantic segmentation, video recognition and image generation, proving its effectiveness.

% \item \textbf{Extensibility to other domains}: In our discussion, we present clues to extend Iwin's interleaved window attention to 1D for large language models and to 3D for video generation, offering a third alternative to conventional 3D full attention and spatial-temporal attention mechanisms, with potential benefits for computational efficiency.
\end{enumerate}

\begin{figure*}[!t]
\centering
\subfloat[]{\includegraphics[width=4.7in]{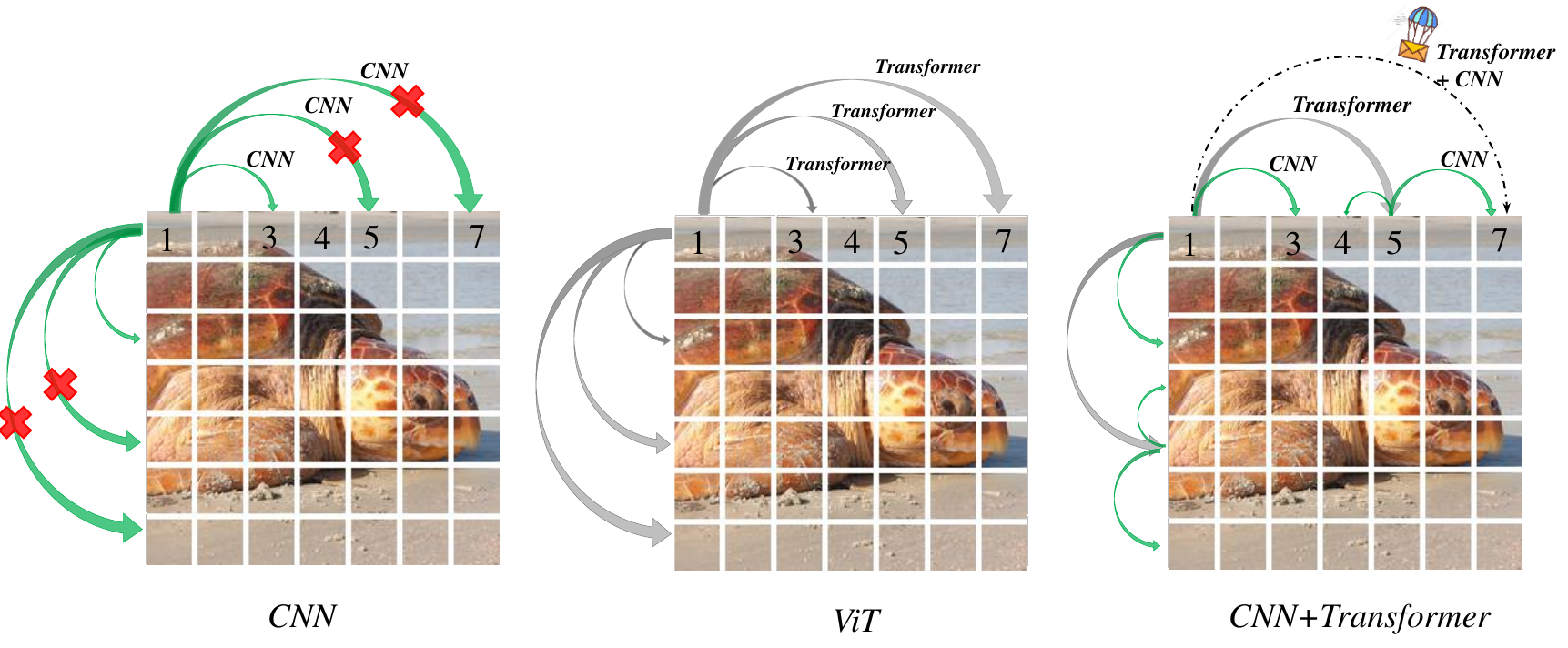}%
\label{fig_first_case}}
\hfil
\subfloat[]{\includegraphics[width=2.4in]{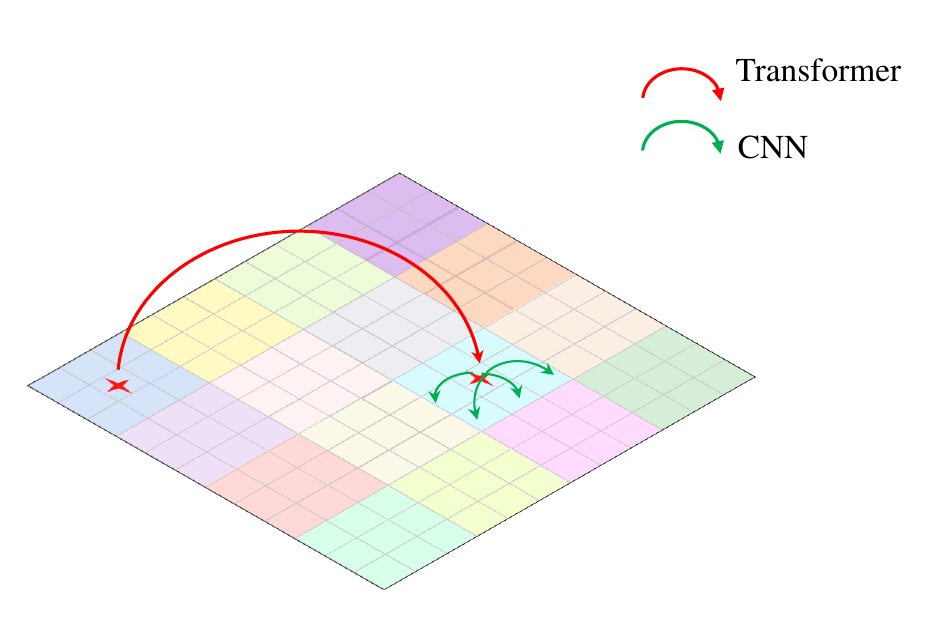}%
\label{fig_second_case}}
\caption{Diagram of the proposed pattern. In (a), token 1 within the CNN can only interact with token 3 nearby and cannot reach token 7 over a long distance. Therefore, CNN is restricted to capturing local features. In contrast, token 1 in the ViT can be associated with any token, enabling the capture of global features but with a quadratic complexity. In the third proposed CNN+Transformer pattern, token 1 first connects with token 5 at a short distance through attention, and token 5 is related to token 7 via convolution. In this way, tokens 1 and 7, despite being far away, communicate indirectly. In (b) shows an intuitive top view of the proposed CNN+Transformer pattern.}
\label{fig:pattern}
\end{figure*}

\section{Related Works}
\label{sec:related}

\subsection{Linear and Sparse Attention}
 Linformer~\cite{wang2020linformer} achieved linear complexity through low-rank factorization of attention matrices, decomposing the $N \times N$ attention matrix into two smaller matrices. Performer~\cite{choromanski2020rethinking} introduced Fast Attention Via positive Orthogonal Random features (FAVOR+), using random feature maps to approximate the attention kernel. BigBird~\cite{zaheer2020big} combined random, window, and global attention patterns to maintain linear complexity. Longformer~\cite{beltagy2020longformer} employed dilated sliding window attention with select global attention tokens. Sparse Transformer~\cite{child2019generating} introduced factorized attention patterns that reduce complexity through structured sparsity. These approaches collectively demonstrate that full global attention is often unnecessary for effective visual representation learning.

\subsection{Hierarchical Vision Transformers}
Hierarchical vision transformers use multi-scale feature representations to improve efficiency while preserving capability. Pyramid Vision Transformer (PVT) \cite{wang2021pyramid} introduces a shrinking pyramid that reduces sequence length in deeper layers via spatial-reduction attention. Swin Transformer \cite{liu2021swin} uses shifted windows, limiting self-attention locally and enabling cross-window connections, reducing complexity from quadratic to linear with respect to image size. CSWin \cite{dong2022cswin} uses cross-shaped self-attention to capture horizontal and vertical dependencies. These designs are effective for dense prediction tasks, where multi-scale features are essential.

\subsection{Hybrid CNN-Transformer Architectures}
Hybrid architectures combine convolutional operations and self-attention to leverage both paradigms. ConViT \cite{d2021convit} uses gated positional self-attention that transitions from convolution to transformer behavior. CoAtNet \cite{dai2021coatnet} unifies depthwise convolutions and self-attention, using convolutions in earlier stages and self-attention later. MobileViT \cite{mehta2021mobilevit} combines local convolutional and global transformer processing for lightweight models. LocalViT \cite{li2021localvit} enhances vision transformers with depth-wise convolutions, adding locality to self-attention layers. These hybrids improve parameter efficiency and performance on smaller datasets while maintaining global modeling capacity for complex vision tasks.

\subsection{Difference With Previous Works} 

Unlike Swin~\cite{liu2021swin}, which needs paired regular- and shifted-window blocks for cross-window exchange, Iwin achieves global connectivity in a single block through collaboration of interleaved window attention and depthwise convolution. The unified module also drops into diffusion models as a direct self-attention replacement without altering text cross-attention placement, a property Swin's two-block design lacks.

Compared with hybrids such as LocalViT~\cite{li2021localvit}, where convolution mainly adds locality, Iwin treats convolution and IWA as interdependent: convolution links token pairs that interleaved attention alone cannot reach while supplying implicit position information that enables position-embedding-free training and direct low-to-high-resolution transfer.

In summary, Iwin is a position-embedding-free hierarchical transformer whose single-block design unifies efficient global exchange, resolution scalability, and straightforward extension from 2D images to 3D video.

\begin{figure*}[!t]
\centering
\includegraphics[width=\textwidth]{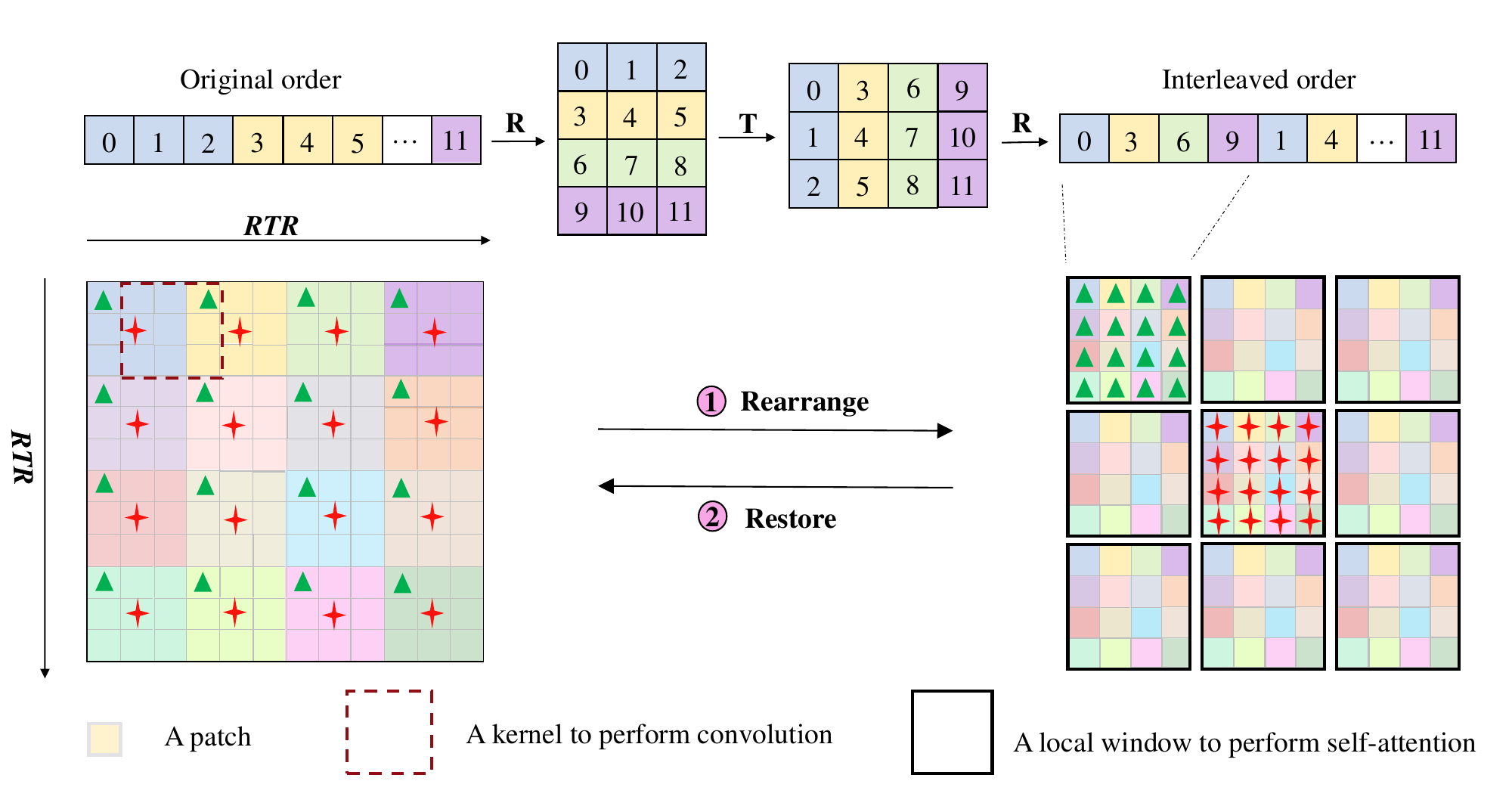}%
\caption{Illustration of Iwin attention. In the left image, the green triangles and red stars representing tokens are connected through convolutions in the original image. In the right image, all green triangles representing tokens are assigned to the same window through the RTR (\textbf{R}eshape-\textbf{T}ranspose-\textbf{R}eshape) operation and window segmentation, executing window attention to establish connections among them. All red stars representing tokens do the same thing. The result is that global convolution and window attention on the interleaved sequence work together to effectively approximate standard global attention, which means that connections are established between any tokens in the original image.}
\label{fig:iwa}
\end{figure*}

\section{Methodology}

\subsection{Overall Architecture}

Iwin Transformer follows a hierarchical architecture similar to Swin Tansformer~\cite{liu2021swin},  progressively reduces spatial resolution while expanding channel dimensions across four stages. Given an input image, Iwin first splits it into non-overlapping patches by a patch splitting module. Each patch is treated as a token or feature,  then architecture processes features through stages $\{S_1, S_2, S_3, S_4\}$ with resolutions $\{H/4, H/8, H/16, H/32\}$ and channels $\{C, 2C, 4C, 8C\}$ respectively. Although pyramid structures were used in this study, the core modules of Iwin can be used in flat structures.

\subsection{Interleaved Window Attention}

Interleaved Window Attention (IWA), as shown in Figure~\ref{fig:iwa}, is the core innovation of the Iwin Transformer. Unlike standard window attention, which evenly divides the feature map into non-overlapping windows, IWA rearranges the feature map before window partition such that tokens from different regions are grouped into the same window for attention computation.

For a feature map $\mathbf{X} \in \mathbb{R}^{H \times W \times C}$, where $H$ and $W$ are the spatial dimensions and $C$ is the channel dimension, IWA operates as follows:

\begin{enumerate}
    \item \textbf{Rearrange}: The input feature map is rearranged such that tokens from different regions are grouped into the same window
    \item \textbf{Attention}: Standard multi-head self-attention is applied to the rearranged tokens
    \item \textbf{Restore}: The tokens are restored to their original spatial arrangement
\end{enumerate}

Supposing window size $M \times M$, we first divide the feature map into non-overlapping $M \times M$ windows containing $H_g \times W_g$ tokens, where $H_g =  H/M $ and $W_g =  W/M $ are the number of windows along the height and width dimensions, respectively.

\subsubsection{Rearrange}
The rearrangement can be expressed as: for a token at position $(i,j)$, its new position in the rearranged feature map is:
\begin{align}
i' &= (i \bmod H_g) \times M + \lfloor i / H_g \rfloor \notag \\
j' &= (j \bmod W_g) \times M + \lfloor j / W_g \rfloor
\end{align}

Luckily, we can elegantly achieve this process through the RTR (\textbf{R}eshape-\textbf{T}ranspose-\textbf{R}eshape) operation as clearly shown in Figure~\ref{fig:iwa}.

Afterwards, the new feature map is evenly divided into $H_g \times W_g$ non-overlapping windows containing  $M \times M$ tokens. 

For a token originally at position $(i,j)$, it will be assigned to the window represented by $(window\_row , window\_col)$:
\begin{align}
window\_row &= \lfloor i' / M \rfloor = \lfloor ((i \bmod H_g) \times M + \lfloor i / H_g \rfloor) / M \rfloor \notag \\
&= \lfloor (i \bmod H_g) + \lfloor i / H_g \rfloor / M \rfloor \notag \\
&= i \bmod H_g \quad \text{(since $\lfloor i / H_g \rfloor / M < 1$)}
\end{align}

Note that $i \bmod H_g$ ranges from $0$ to $H_g-1$, which exactly corresponds to the row index of the window in the grid of windows. Similarly, $window\_col = j \bmod W_g$ ranges from $0$ to $W_g-1$, corresponding to the column index of the window.

This ensures that tokens with the same $(i \bmod H_g, j \bmod W_g)$ are grouped into the same window for attention computation. That means, if two tokens at $(i_1,j_1)$ and $(i_2,j_2)$ are in the same window, they must satisfy:
\begin{equation}
    i_1 \bmod H_g = i_2 \bmod H_g \text{ and } j_1 \bmod W_g = j_2 \bmod W_g
\end{equation}

\subsubsection{Self-Attention}

Within each window, we apply the standard self-attention mechanism~\cite{vaswani2017attention}:

\begin{align}
\mathbf{Q} = \mathbf{X}\mathbf{W}_Q  \quad
\mathbf{K} = \mathbf{X}\mathbf{W}_K \quad
\mathbf{V} = \mathbf{X}\mathbf{W}_V \notag \\
\text{Attention}(\mathbf{Q}, \mathbf{K}, \mathbf{V}) = \text{Softmax}\left(\frac{\mathbf{Q}\mathbf{K}^T}{\sqrt{d_k}}\right)\mathbf{V}
\end{align}

where $\mathbf{W}_Q$, $\mathbf{W}_K$, and $\mathbf{W}_V$ are learnable projection matrices, and $d_k$ is the dimension of the key vectors.

\subsubsection{Restore}

Finally, the tokens are restored to their original spatial arrangement using the inverse RTR, which is also a Reshape-Transpose-Reshape operation as shown in Figure~\ref{fig:iwa} to realize $(i', j') \mapsto (i, j)$:
% $RTR^{-1}$
\begin{align}
i &= (i' \bmod M) \times H_g + \lfloor i' / M \rfloor \notag \\
j &= (j' \bmod M) \times W_g + \lfloor j' / M \rfloor
\end{align}

\subsection{Depthwise Convolution}

Depthwise Convolution (DWConv)~\cite{chollet2017xception} is used to help build missing relationships of certain tokens that are not in the same attention window and provide implicit positional information by the way~\cite{islam2020much}. This makes Iwin get rid of positional encodings and significantly enhances scalability, a beneficial byproduct.

\subsection{Downsampling Layer}

Iwin Transformer employs standard convolution to progressively reduce spatial resolution and increase channel dimensions, following the hierarchical design principle ~\cite{he2016deep} common in vision backbones. 

\begin{equation}
\mathcal{D}(\mathbf{X}) = \text{Conv}_{3 \times 3, \text{stride}=2}(\mathbf{X})
\end{equation}

We tested four downsampling methods as shown in Table~\ref{tab:ablations}: average pooling, patch merging, standard convolution, and depthwise convolution. They all worked very well, with only a 0.2\% performance difference between them. We chose standard convolution, which had the highest accuracy.

\subsection{Iwin Transformer Block}
The Iwin Transformer block can be formulated as:
\begin{align}
\mathbf{X}' &=\text{LayerNorm}(\mathbf{X})  \notag \\
\mathbf{X}'' &= \mathbf{X} + \text{IW-MSA}(\mathbf{X}') + \text{DWConv}(\mathbf{X}') \notag \\
\mathbf{X}''' &= \mathbf{X}'' + \text{MLP}(\text{LayerNorm}(\mathbf{X}''))
\end{align}
where IW-MSA is Interleaved Window Multi-Head Self-Attention. The computation cost of the unified module combining IW-MSA and DWConv as follows:
\begin{align}
\mathcal{O}_{Iwin} &= \underbrace{\frac{HW}{M^2} \times 3M^2C^2}_{\text{QKV projection}} + \underbrace{2\frac{HW}{M^2} \times M^4C}_{\text{Attention computation}} \notag \\
&\quad + \underbrace{\frac{HW}{M^2} \times M^2C^2}_{\text{Output projection}} + \underbrace{HW \times C \times k^2}_{\text{Convolution}} \notag \\
&=4HWC^2 + (2M^2+k^2)HWC
\label{eq:cost}
\end{align}
Compared to Swin
\begin{align}
\mathcal{O}_{Swin}=4HWC^2 + 2M^2HWC
\end{align}

Although Iwin adds $k^2HWC$ FLOPs over Swin, the overhead is modest when $M \gg k$ and is offset by architectural gains. First, a single Iwin block replaces two consecutive Swin blocks for cross-window interaction and can be applied to diffusion models as a standalone module. Second, depthwise convolution injects implicit spatial structure, eliminating explicit position embeddings and allowing models pre-trained at $224^2$ to be directly fine-tuned to $384^2$--$1024^2$ without bias interpolation.

\begin{algorithm}[H]
  \caption{\small Pseudocode of \textbf{Rearrange} and \textbf{Restore} Operations in a PyTorch-like style.}
  \label{alg:code}
  \definecolor{codeblue}{rgb}{0.25,0.6,0.6}
  \definecolor{codekw}{rgb}{0.0, 0.0, 0.0}
  \lstset{
    backgroundcolor=\color{white},
    basicstyle=\fontsize{6.95pt}{6.95pt}\ttfamily\selectfont,
    columns=fullflexible,
    numbers=none,
    breaklines=true,
    captionpos=b,
    commentstyle=\fontsize{6.95pt}{6.95pt}\color{codeblue},
    keywordstyle=\fontsize{6.95pt}{6.95pt}\color{codekw},
  }
  \begin{lstlisting}[language=python]

    # H_num_win:  H // window_size, which means the number of windows along H
    # W_num_win:  w // window_size, which means the number of windows along W

    def rearrange(x, H_num_win, W_num_win):
        B, H, W, C = x.shape
    
        x = x.reshape(B, -1, H_num_win, W, C).transpose(1, 2)
        x = x.reshape(B, -1, W, C)
        
        x = x.reshape(B, H, -1, W_num_win, C).transpose(2, 3)
        x = x.reshape(B, H, -1, C)
        return x


    def restore(x, H_num_win, W_num_win):
        B, H, W, C = x.shape
    
        x = x.reshape(B, H, W_num_win, -1, C).transpose(2, 3)
        x = x.reshape(B, H, -1, C)
    
        x = x.reshape(B, H_num_win, -1, W, C).transpose(1, 2)
        x = x.reshape(B, -1, W, C)
        return x

  \end{lstlisting}
\label{fig:code}
\end{algorithm}

\subsection{Global Information Exchange}

We believe that global information exchange is achieved when there is a path in the feature map along which information can flow from one position $(i_1,j_1)$ to another $(i_2,j_2)$.

\begin{lemma}[Modular Property of Interleaved Window Attention]
\label{lemma:iwa_modular}
In interleaved window attention, tokens at positions $(i_1,j_1)$ and $(i_2,j_2)$ are in the same attention window if and only if:
\[i_1 \bmod H_g = i_2 \bmod H_g \text{ and } j_1 \bmod W_g = j_2 \bmod W_g\]
\end{lemma}

\begin{lemma}[Locality of Depthwise Separable Convolution]
\label{lemma:dsc_locality}
For depthwise separable convolution with kernel size $K \times K$, tokens at positions $(i_1,j_1)$ and $(i_2,j_2)$ can directly exchange information if and only if:
\[|i_1 - i_2| \leq K \text{ and } |j_1 - j_2| \leq K\]
\end{lemma}

Based on these lemmas, we prove the following theorem:

\begin{theorem}[Global Information Exchange Condition]
\label{thm:global_info_exchange}
If $K M\geq \max(H,W)$, where $K$ is kernel size and $M$ is window size, then the combination of interleaved window attention and depthwise separable convolution in the Iwin Transformer block enables information exchange between any two positions $(i_1,j_1)$ and $(i_2,j_2)$ in the feature map.
\end{theorem}

Consider arbitrary two positions $(i_1,j_1)$ and $(i_2,j_2)$ in the feature map. We need to prove that there exists a path for information to flow from $(i_1,j_1)$ to $(i_2,j_2)$.

We discuss three cases:

\textbf{Case 1}: $(i_1 \bmod H_g = i_2 \bmod H_g)$ and $(j_1 \bmod W_g = j_2 \bmod W_g)$

In this case, according to Lemma~\ref{lemma:iwa_modular}, positions $(i_1,j_1)$ and $(i_2,j_2)$ are in the same attention window, so they can directly exchange information through the attention mechanism.

\textbf{Case 2}: $(|i_1 - i_2| \leq K \quad and \quad |j_1 - j_2| \leq K)$

In this case, according to Lemma~\ref{lemma:dsc_locality}, positions $(i_1,j_1)$ and $(i_2,j_2)$ are in the same convolution kernel, so they can directly exchange information through the convolution mechanism.

\textbf{Case 3}: Otherwise (i.e., when $(i_1,j_1)$ and $(i_2,j_2)$ are not in the same attention window and convolution kernel)

In this case, we need to find an intermediate position $(i_3,j_3)$ to bridge  $(i_1,j_1)$ and $(i_2,j_2)$. 

We construct such a position $(i_3,j_3)$ as follows:
\begin{align}
i_3 &= (i_1 \bmod H_g) + H_g \cdot \lfloor i_2 / H_g \rfloor \notag \\
j_3 &= (j_1 \bmod W_g) + W_g \cdot \lfloor j_2 / W_g \rfloor
\end{align}

Now we have
\begin{align}
i_3 \bmod H_g &= ((i_1 \bmod H_g) + H_g \cdot \lfloor i_2 / H_g \rfloor) \bmod H_g \notag \\
&= (i_1 \bmod H_g) \bmod H_g + (H_g \cdot \lfloor i_2 / H_g \rfloor) \bmod H_g \notag \\
&= (i_1 \bmod H_g) + 0 \notag \\
&= i_1 \bmod H_g
\end{align}

Similarly, $j_3 \bmod W_g = j_1 \bmod W_g$. This means that positions $(i_1,j_1)$ and $(i_3,j_3)$ are in the same attention window according to Lemma~\ref{lemma:iwa_modular}.

Now we check:
\begin{align}
|i_2 - i_3| &= |i_2 - ((i_1 \bmod H_g) + H_g \cdot \lfloor i_2 / H_g \rfloor)| \notag \\
&= |i_2 - (i_1 \bmod H_g) - H_g \cdot \lfloor i_2 / H_g \rfloor| \notag \\
&= |i_2 \bmod H_g - i_1 \bmod H_g| \notag \\
&\leq H_g - 1 \notag \\
& = H/M  - 1 \notag \\
&< H/M 
\end{align}

Similarly, $|j_2 - j_3| < W/M $.

When $KM \geq \max(H,W)$, we have $|i_2 - i_3| \leq K$ and $|j_2 - j_3| \leq K$, so positions $(i_3,j_3)$ and $(i_2,j_2)$ can directly exchange information through depthwise separable convolution. 

Therefore, the Iwin Transformer block enables information exchange between any two positions in the feature map when $K M \geq \max(H,W)$. There always exists $(i_3,j_3)$ such that
\begin{align}
i_1 \bmod H_g &= i_3 \bmod H_g  \notag \\
j_1 \bmod W_g &= j_3 \bmod W_g \notag \\
|i_2 - i_3| &\leq K \notag \\
|j_2 - j_3| &\leq K
\end{align}

This means that positions $(i_1,j_1)$ and $(i_3,j_3)$ are connected through interleaved window attention, while positions $(i_3,j_3)$ and $(i_2,j_2)$ are connected through depthwise separable convolution, and $(i_1,j_1)$ and $(i_2,j_2)$  established a connection through $(i_3,j_3)$ as an intermediary bridge.

$KM \geq \max(H, W)$ provides a \emph{design guideline} rather than a strict prescription. In a hierarchical pyramid, each stage has a different spatial resolution. At first, we followed the rule, assigning different convolution kernel sizes to each stage. Our ablation (Table~\ref{tab:ablations}) shows that per-stage kernels $\{7,5,3\}$ underperform a uniform kernel (e.g., $\{3,3,3\}$), mainly due to imbalanced stage capacity~\cite{tan2019efficientnet}, not because the theory is invalid. We believe stacked blocks and downsampling enlarge the effective receptive field (ERF)~\cite{luo2016understanding}, so a fixed small kernel can still satisfy $K_{\mathrm{ERF}} M \geq \max(H, W)$ at sufficient depth. When it applies to flat architectures. In FlashDiT (Sec.~\ref{sec:imgen}), to meet $KM \geq \max(H, W)$, the larger window yields better generation quality, proving its correctness.

\section{Experiments}

We conduct experiments on ImageNet-1K image classification~\cite{deng2009imagenet}, ADE20K semantic segmentation~\cite{zhou2019semantic}, Kinetics-400~\cite{kay2017kinetics} video recognization and class-conditional image generation. And finally we ablate the important design elements of Iwin Transformer.

\subsection{Image Classification on ImageNet-1K}

\paragraph{\textbf{Settings}}
For image classification, we benchmark the proposed Iwin Transformer on ImageNet-1K~\cite{deng2009imagenet}, which contains 1.28M training images and 50K validation images from 1,000 classes. Experimental settings closely follow those of Swin Transformer~\cite{liu2021swin}. We report top-1 accuracy on a single crop under two scenarios: training from scratch on ImageNet-1K and pre-training on ImageNet-22K, which contains 14.2M images and 22K classes, followed by fine-tuning on ImageNet-1K.

\emph{\textbf{From scratch on ImageNet-1K}}. We use an AdamW optimizer~\cite{kingma2017adam} for 300 epochs with a 20-epoch linear warmup and a cosine learning rate decay. We set the batch size to 512, weight decay to 0.05, and use most of the augmentation strategies from DeiT~\cite{touvron2020deit}. The initial learning rate is set to 0.0005, and the drop path rates are 0.2, 0.3, and 0.5 for Iwin-T, Iwin-S, and Iwin-B, respectively.

\emph{\textbf{Pre-training on ImageNet-22K}}. Training lasts for 90 epochs with a 5-epoch warmup. We use a batch size of 4096, an initial learning rate of $1.25e-4$, and a weight decay of 0.05. The pre-trained models are then fine-tuned on ImageNet-1K for 10~30 epochs at a $224\times224$ resolution, using a batch size of 1024, a constant learning rate of $2e-05$, and a weight decay of $1e-8$.

\emph{\textbf{Cross-Resolution Fine-tuning}}. A key advantage of Iwin Transformer is it’s simple to transfer to higher resolutions. Unlike conventional practices that require staged fine-tuning or interpolate absolute/relative position biases, our models pre-trained at $224^2$ resolution can be directly fine-tuned on higher resolutions such as $384^2$, $512^2$, and $1024^2$. The only thing need to do is to change the hyperparameter window size. We adjust the window size to match the input resolution—7 for 224, 12 for 384, and 16 for 512 and 1024—while keeping the architectures unchanged. The fine-tuning process runs for 30 epochs with a constant learning rate of $2e-05$, a weight decay of $1e-8$, and a warm-up of 5 to 10 epochs.

\paragraph{\textbf{Results}} Table \ref{tab:imagenet1k} presents comprehensive results of the Iwin Transformer's competitive performance and resolution adaptability.

\begin{table}[t]
  \centering
  \caption{Classification accuracy on ImageNet-1K. 
  Throughput (img/s) is measured on a V100 GPU with PyTorch. For Iwin, each entry is \textit{vanilla\,/\,Flash Attention} (e.g., 729/755); other methods report vanilla attention only.}
  \label{tab:imagenet1k}
  \scalebox{0.85}{
  \begin{tabular}{c|cccc|c}
  \toprule
  \multirow{2}{*}{Method}             & Image Size   & Param         & FLOPs          & Throughput     & Top-1 Acc  \\ 
                                      & (px)         & (M)           & (G)            & (img/s)        & (\%)       \\   \midrule
  
  \multicolumn{6}{c}{\textbf{(a) ImageNet-1K trained models}}                                                                                      \\   \midrule
  DeiT-Small/16\cite{touvron2020deit}            & 224$^2$          & 22.0          & 4.6            & 406            & 79.9        \\
  % T2T-ViT-14\cite{T2T}                & 224$^2$          & 22.0          & 5.2            & -              & 81.5        \\
  PVT-Small\cite{PVT_v1}              & 224$^2$          & 24.5          & 3.8            & 794            & 79.8        \\
  % Twins-S\cite{chu2021twins}                 & 224$^2$          & 24.0          & 2.9            & 979            & 81.7        \\ 
  PVT-v2-B2\cite{PVT_v2}              & 224$^2$          & 25.4          & 4.0            & 664            & 82.0        \\
  % PoolFormer-S36\cite{metaformer}     & 224$^2$          & 31.0          & 5.1            & 764            & 81.4        \\ 
  Swin-T\cite{liu2021swin}                   & 224$^2$          & 29.0          & 4.5            & 758            & 81.3        \\
  ConvNeXt-T~\cite{liu2022convnet} & 224$^2$ & 29 & 4.5 & 775 & 82.1 \\ 
  {\bf Iwin-T(ours)}                & 224$^2$          & 30.2    & 4.7      & 729/755     & {\bf 82.0}  \\  
  
  \midrule                            
  
  % T2T-ViT-19\cite{T2T}                & 224$^2$          & 39.2          & 8.9            & -              & 81.9        \\

  PVT-Medium\cite{PVT_v1}             & 224$^2$          & 44.2          & 6.7            & 511            & 81.2        \\
  % Twins-B\cite{chu2021twins}                 & 224$^2$          & 56.0          & 8.6            & 433            & 83.2        \\
  PVT-v2-B3\cite{PVT_v2}              & 224$^2$          & 45.2          & 6.9            & 443            & 83.2        \\
  % PoolFormer-M36\cite{metaformer}     & 224$^2$          & 56.0          & 9.0            & 494            & 82.1        \\ 
  Swin-S\cite{liu2021swin}                   & 224$^2$          & 50.0          & 8.7            & 437            & 83.0        \\
  ConvNeXt-S~\cite{liu2022convnet} & 224$^2$ & 50 & 8.7 & 447 & 83.1 \\ 
  {\bf Iwin-S(ours)}                & 224$^2$          & 51.6    & 9.0      &  410/435      & {\bf 83.4}  \\  
  {\bf Iwin-S(ours)}                & 384$^2$          & 51.6    & 27.7     & 142/151     & {\bf 84.3} \\
  {\bf Iwin-S(ours)}                & 512$^2$          & 51.6    & 52.0      &  78/86     & {\bf 84.4}  \\  
  {\bf Iwin-S(ours)}                & 1024$^2$          & 51.6    & 207.9      & 20/23      & {\bf 83.8}  \\  
  
  \midrule                      
  
  DeiT-Base/16\cite{touvron2020deit}             & 224$^2$          & 86.6          & 17.6           & 273            & 81.8        \\
  % T2T-ViT-24\cite{T2T}                & 224$^2$          & 64.1          & 14.1           & -              & 82.3        \\
  PVT-Large\cite{PVT_v1}              & 224$^2$          & 61.4          & 9.8            & 357            & 81.7        \\
  % Twins-L\cite{chu2021twins}                 & 224$^2$          & 99.2          & 15.1           & 271            & 83.7        \\
  PVT-v2-B4\cite{PVT_v2}              & 224$^2$          & 62.6          & 10.1           & 298            & 83.6        \\
  % PoolFormer-M48\cite{metaformer}     & 224$^2$          & 73.0          & 11.8           & 337            & 82.5        \\ 
  Swin-B\cite{liu2021swin}                   & 224$^2$          & 88.0          & 15.4           & 287            & 83.3        \\
  Swin-B\cite{liu2021swin}                   & 384$^2$          & 88.0          & 47.0           & 85            & 84.5        \\
  ConvNeXt-B~\cite{liu2022convnet} & 224$^2$ & 89 & 15.4 & 292& 83.8 \\ 
  ConvNeXt-B~\cite{liu2022convnet} & 384$^2$ & 89 & 45.0 & 96 & 85.1 \\
  
  {\bf Iwin-B(ours)}                & 224$^2$          & 91.2    & 15.9     &   271/286    & {\bf 83.5}  \\  
  {\bf Iwin-B(ours)}                & 384$^2$          &91.2    & 48.3     &     78/87  & {\bf 84.9}  \\
  {\bf Iwin-B(ours)}                & 512$^2$          & 91.3    & 89.5     &    51/55   & {\bf 85.1}  \\
  {\bf Iwin-B(ours)}                & 1024$^2$          & 91.3    & 358.2     &   12/14    & {\bf 85.0}  \\
  \midrule
  \multicolumn{6}{c}{\textbf{(b) ImageNet-22K pre-trained models}}  \\  
  \midrule

   R-101x3~\cite{kolesnikov2020bigtransferbitgeneral} & 384$^2$ & 388 & 204.6 & - & 84.4 \\
   R-152x4~\cite{kolesnikov2020bigtransferbitgeneral} & 480$^2$ & 937 & 840.5 & - & 85.4 \\
   ViT-B/16~\cite{dosovitskiy2021imageworth16x16words} & 384$^2$ & 86 & 55.4 & 86 & 84.0 \\
   ViT-L/16~\cite{dosovitskiy2021imageworth16x16words} & 384$^2$ & 307 & 190.7 & 27 & 85.2 \\
   Swin-B~\cite{liu2021swin} & 224$^2$ & 88 & 15.4 & 287 & 85.2 \\
   Swin-B~\cite{liu2021swin} & 384$^2$ & 88 & 47.0 & 85 & 86.4 \\
   ConvNeXt-B~\cite{liu2022convnet} & 224$^2$ & 89  & 15.4 & 292 & 85.8 \\
   ConvNeXt-B~\cite{liu2022convnet}  & 384$^2$ & 89  & 45.1 & 96  & 86.8 \\
   % Swin-L & 384$^2$ & 197 & 103.9 & 42.1 & 87.3  \\
  {\bf Iwin-B(ours)}                & 224$^2$          & 91.2    & 15.9     &    271/286  & {\bf 85.5}  \\  
  {\bf Iwin-B(ours)}                & 384$^2$          &91.2    & 48.3     &      78/87  & {\bf 86.6}  \\
  {\bf Iwin-B(ours)}                & 512$^2$          & 91.2    & 89.5     &     51/55  & {\bf 86.1}  \\
  {\bf Iwin-B(ours)}                & 1024$^2$          & 91.2    & 358.2     &   12/14  & {\bf 85.6}  \\
   Swin-L~\cite{liu2021swin} & 224$^2$ & 197 & 34.5 & 145 & 86.3 \\
   Swin-L~\cite{liu2021swin} & 384$^2$ & 197 & 103.9 & 46 & 87.3 \\
   ConvNeXt-L~\cite{liu2022convnet} & 224$^2$ & 198 & 34.4 & 147 & 86.6\\
   ConvNeXt-L~\cite{liu2022convnet} & 384$^2$ & 198 & 101.0& 50  & 87.5  \\
   {\bf Iwin-L(ours)}                & 224$^2$          & 204.3    & 35.4     &   138/147  & {\bf 86.4}  \\
   {\bf Iwin-L(ours)}                & 384$^2$          & 204.3    & 106.6     &   43/48  & {\bf 87.4}  \\
   
  \bottomrule
  \end{tabular}
  }
  \vspace{-0.3cm}
\end{table}

\begin{figure}[ht]
\centering
\begin{minipage}[b]{0.48\textwidth}
\centering
{\includegraphics[width=\textwidth]{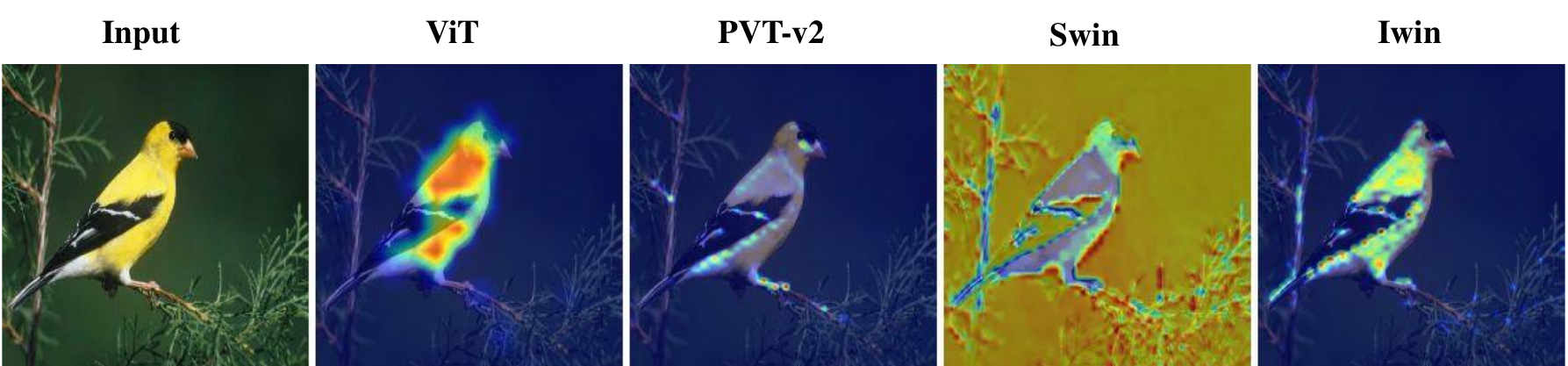}} \\
\includegraphics[width=\textwidth]{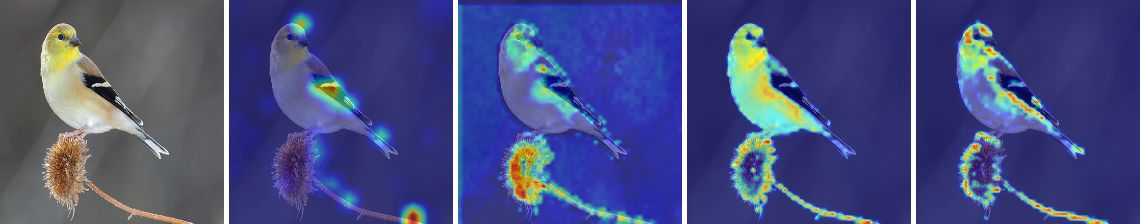} \\
\includegraphics[width=\textwidth]{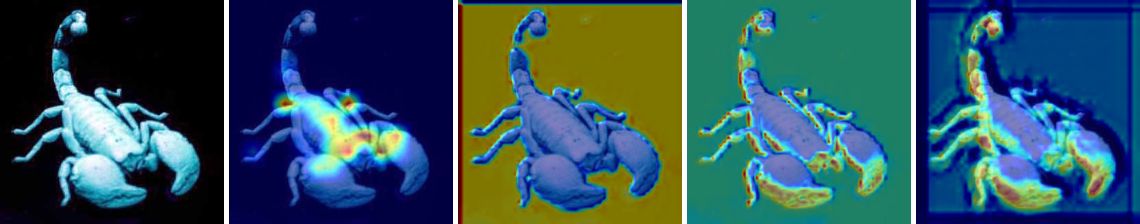} \\
\includegraphics[width=\textwidth]{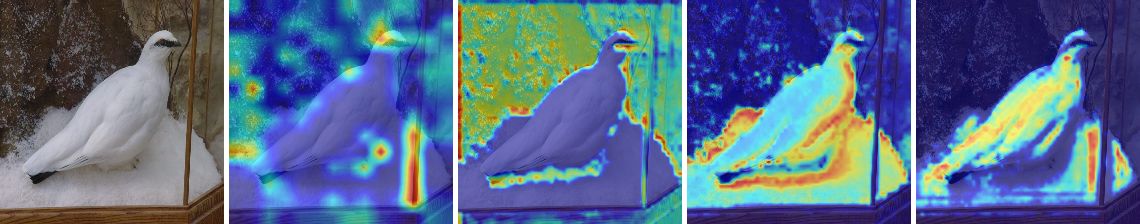} \\
\includegraphics[width=\textwidth]{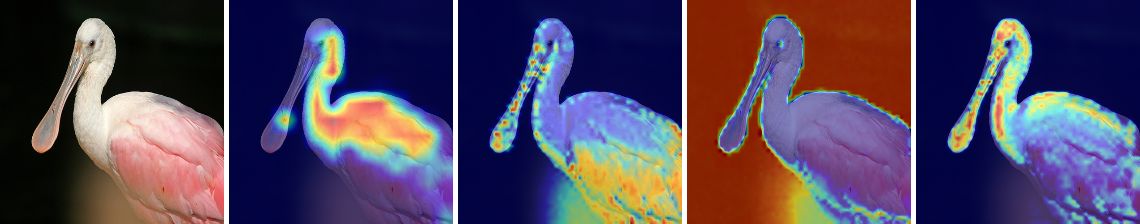}
\end{minipage}

\caption{The visualization of heatmap. The left column shows input images, while subsequent columns show results from native VIT, PVTv2, Swin, and Iwin (our method). Results demonstrate that Iwin effectively concentrates activation on target objects.}
\label{fig:heatmap}
\end{figure}

\emph{\textbf{ImageNet-1K}}. For models trained from scratch on ImageNet-1K at a resolution of 224$\times$224, Iwin Transformer variants consistently demonstrate competitive Top-1 accuracy. Iwin-T achieves \textbf{82.0\%}, matching ConvNeXt-T (82.1\%) and surpassing Swin-T (81.3\%) by \textbf{0.7\%}. The deeper variant, Iwin-S, reaches highest \textbf{83.4\%}, outperforming Swin-S (83.0\%) by \textbf{0.4\%} and ConvNeXt-S (83.1\%) by \textbf{0.3\%}, while maintaining similar parameters (51.6M vs. 50.0M) and FLOPs (9.0G vs. 8.7G) compared to both Swin-T and ConvNeXt-T. By increasing the embedding dimension, Iwin-B achieves 83.5\%, which is close to the 83.6\% of PVT-v2-B4 and falls between the 83.3\% of Swin-B and the 83.8\% of ConvNeXt-B.

\emph{\textbf{ImageNet-22K}}. By leveraging large-scale ImageNet-22K pre-training, Iwin significantly enhances its performance. The pre-trained Iwin-B achieves \textbf{85.5\%} Top-1 accuracy at a resolution of 224$\times$224. When fine-tuned to a resolution of 384$\times$384, it reaches \textbf{86.6\%}, surpassing Swin-B (86.4\%) by \textbf{0.2\%} and closely aligning with the carefully configured ConvNeXt-B (86.8\%). The largest variant, Iwin-L, demonstrates cutting-edge performance with \textbf{86.4\%} at 224$\times$224 and an impressive \textbf{87.4\%} at 384$\times$384, competing favorably with top-tier models like Swin-L (87.3\%) and ConvNeXt-L (87.5\%).

\emph{\textbf{Cross-Resolution Fine-tuning}}. Iwin models have the capability to fine-tune directly from a $224^2$ pre-trained state to higher resolutions because there are no positional embedding limits. Iwin-S can be fine-tuned to achieve Top-1 accuracies of \textbf{84.3\%} at $384^2$ (a \textbf{0.9\%} increase from its $224^2$ baseline of 83.4\%) and \textbf{84.4\%} at $512^2$ (a \textbf{1.0\%} gain). Similarly, Iwin-B reaches \textbf{84.9\%} at $384^2$ (a \textbf{1.4\%} improvement from its $224^2$ baseline of 83.5\%). Furthermore, Iwin-B demonstrates robust performance at even higher resolutions, achieving \textbf{85.1\%} at $512^2$ (a \textbf{1.6\%} gain). For models pre-trained on ImageNet-22K, this advantage is equally pronounced. Iwin-B fine-tunes from $224^2$ to $384^2$ (86.6\%, a \textbf{1.1\%} gain from its $224^2$ baseline of 85.5\%), outperforming Swin-B ($86.4\%$) by \textbf{0.2\%} at $384^2$. It continues to perform strongly at $512^2$ (86.1\%) and $1024^2$ (85.6\%). Meanwhile, Iwin-L transitions from $224^2$ to $384^2$ (\textbf{87.4\%}, a 1.0\% gain), standing with Swin-L ($87.3\%$) and ConvNeXt-L ($87.5\%$).

\emph{\textbf{Throughput}}. Iwin reports throughput under vanilla attention\,/\,Flash Attention~\cite{dao2022flashattention}; other models use vanilla only. That's because Iwin can enjoy the efficient attention implementations, whereas Swin cannot because of the irregular mask. At $224^2$, vanilla Iwin trails Swin (e.g., Iwin-T 729 vs. 758) because extra depthwise convolution is memory-bound and its impact is large at a small scale; with Flash enabled, Iwin-T/S/B approach Swin (755/758, 435/437, 286/287 img/s). As resolution and model capacity increase, attention (compute-bound) dominates FLOPs and memory pressure, and Flash Attention yields larger gains. For Iwin-B, the flash speedup grows from +5.5\% at $224^2$ to +11.5\% at $384^2$, +7.8\% at $512^2$, and +16.7\% at $1024^2$; at $384^2$, Iwin-B flash throughput (87\,img/s) exceeds Swin-B (85\,img/s).

\subsection{Semantic Segmentation on ADE20K}

\paragraph{\textbf{Settings}}  We evaluate the Iwin backbone on the ADE20K~\cite{zhou2019semantic} semantic segmentation benchmark using UperNet~\cite{xiao2018upernet} with the MMSegmentation~\cite{mmseg2020} toolbox. The experimental settings follow Swin~\cite{liu2021swin}. Training is performed for 160K iterations with a total batch size of 16 (2 images per GPU across 8 GPUs). 

\begin{table}[ht]
  \centering
   \caption{Results for ADE20K semantic segmentation task. FLOPs are measured with the input size of 512$\times$2048.}
  \resizebox{\columnwidth}{!}{
  \begin{tabular}{l|ccc|ccc}
  \toprule
  \multirow{2.0}{*}{Backbone}     & \multicolumn{3}{c|}{Semantic FPN 80k}   & \multicolumn{3}{c}{UperNet 160k}          \\ \cmidrule(l){2-7}
                                              & Param(M)    & FLOPs(G)   & mIoU(\%)      & Param(M)  &   FLOPs(G)     &    mIoU(\%)        \\   \midrule
  ResNet50\cite{he2016deep}                       & 28.5        & 183        & 36.7          & -            & -           & -                     \\
 
  PVTv2 B2\cite{PVT_v1}                      & 29.1        & 165        & 45.2          & -            & -           & -                     \\
   ConvNeXt-T\cite{liu2022convnet}                   & -           & -          & -             & 60.0         & 939         & 46.0                  \\
  Swin-T\cite{liu2021swin}                           & -        & -        & -          & 59.9         & 945         & 44.5   \\
  {\bf Iwin-T(ours)}                        & -  & -  & -   & {\bf 61.9} & {\bf 946} & {\bf 44.7}   \\ \midrule
  
  ResNet101\cite{he2016deep}                      & 47.5        & 260        & 38.8          & -         & -        & -                  \\
 
  PVTv2 B3\cite{PVT_v1}                     & 49.0        & 224        & 47.3          & -            & -           & -                     \\
  ConvNeXt-S\cite{liu2022convnet}                   & -           & -          & -             & 82.0         & 1027        & 48.7                  \\
  Swin-S\cite{liu2021swin}                           & -        & -        & -          & 81.3         & 1038        & 47.6               \\

  {\bf Iwin-S(ours)}                        & -  & - & -   & {\bf 83.2}   & {\bf 1038}  & {\bf 47.5}   \\   \midrule
  
  ResNeXt101-64$\times$4d\cite{ResNeXt}       & 86.4        & -          & 40.2          & -            & -           & -                     \\
  
  PVTv2 B4\cite{PVT_v1}                      & 66.3        & 285        & 47.9          & -            & -           & -                     \\
  ConvNeXt-B\cite{liu2022convnet}                   & -           & -          & -             & 122.0        & 1170        & 49.1                  \\
  Swin-B\cite{liu2021swin}                          & -        & -        & -          & 121.0        & 1188        & 48.1               \\  

  {\bf Iwin-B(ours)}                        &  -  & -  & -    & {\bf 124.8} & {\bf 1189} & {\bf 48.9}   \\ 
  \bottomrule
  \end{tabular}}

  \label{tab:ade20K}
\end{table}

\paragraph{\textbf{Results}} As shown in Table~\ref{tab:ade20K}, Iwin-B achieves a 48.9\% mIoU, surpassing Swin-B's 48.1\% mIoU by 0.8\%, while maintaining nearly identical FLOPs (1189G vs 1188G) and parameters (124.8M vs 121.0M). This performance is close to the leading ConvNeXt-B, which records a 49.1\% mIoU with a similar computational cost. For smaller models, Iwin-T achieves a 44.7\% mIoU, slightly exceeding Swin-T's 44.5\% mIoU with comparable FLOPs. These results demonstrate Iwin's effectiveness and competitiveness in semantic segmentation tasks.

\begin{figure}[!t]
\centering
{
\includegraphics[width=3.5 in]{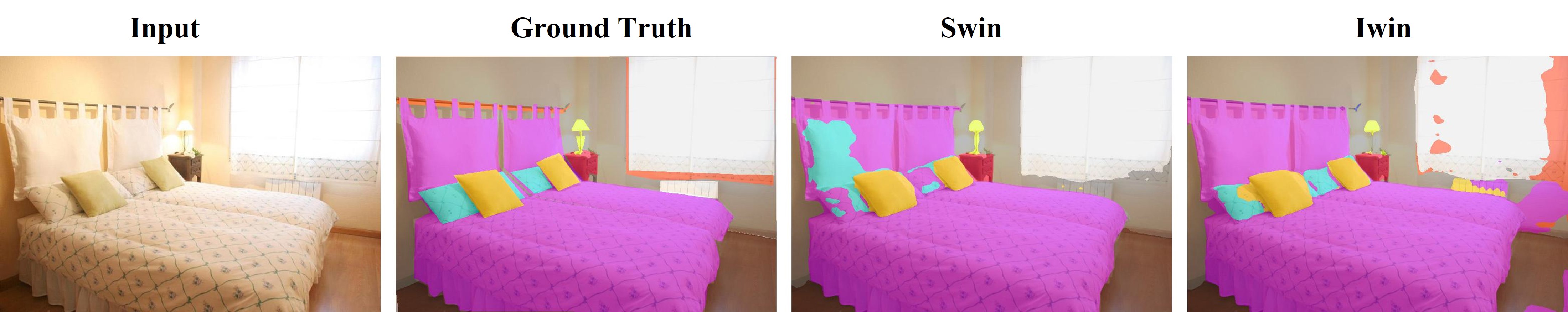}
\\
\includegraphics[width=3.5 in]{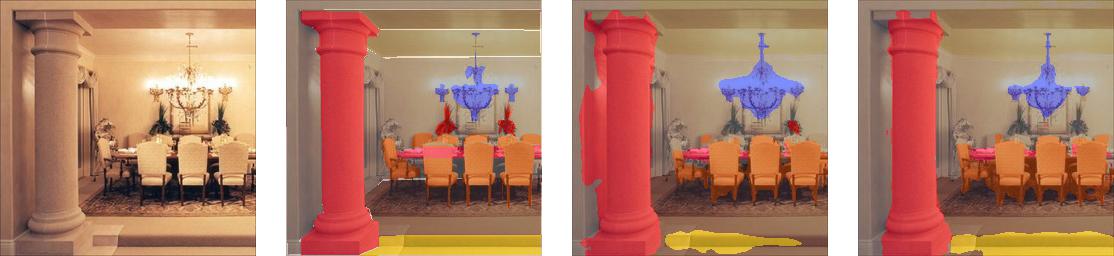}% 
\\
\includegraphics[width=3.5 in]{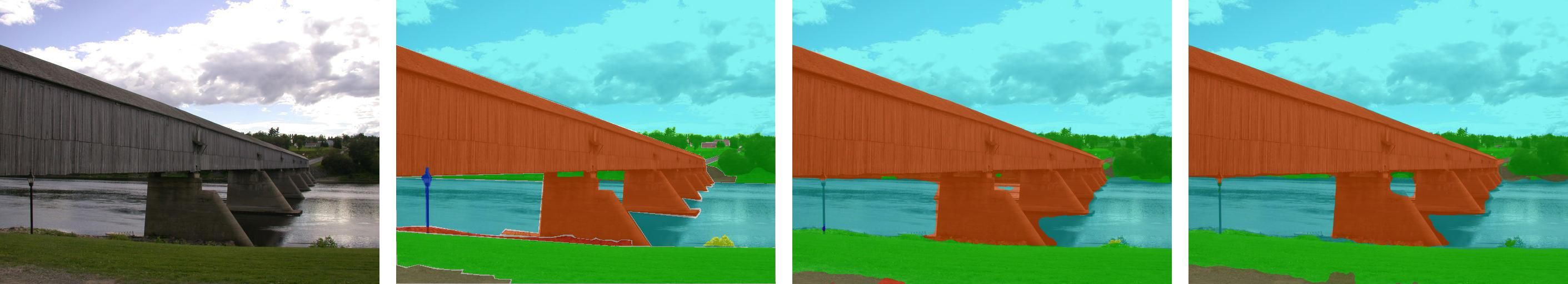}% 
\\
\includegraphics[width=3.5 in]{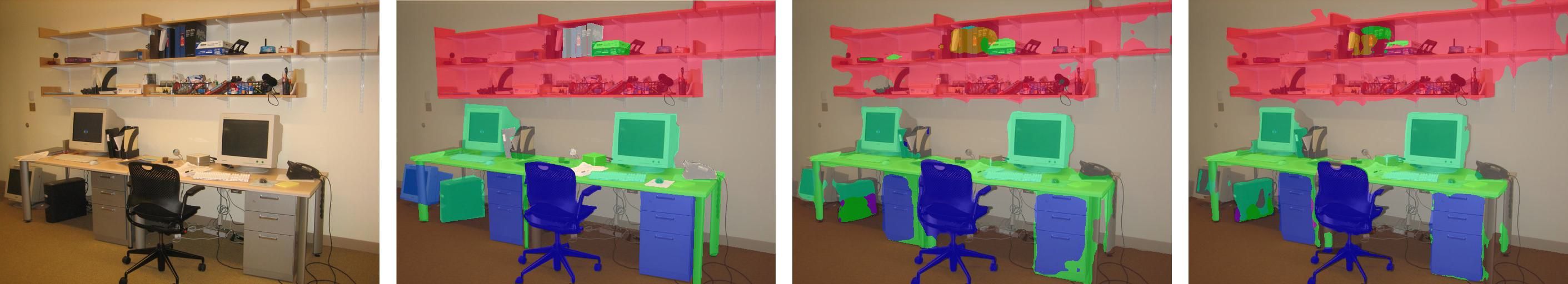}%
\\
\includegraphics[width=3.5 in]{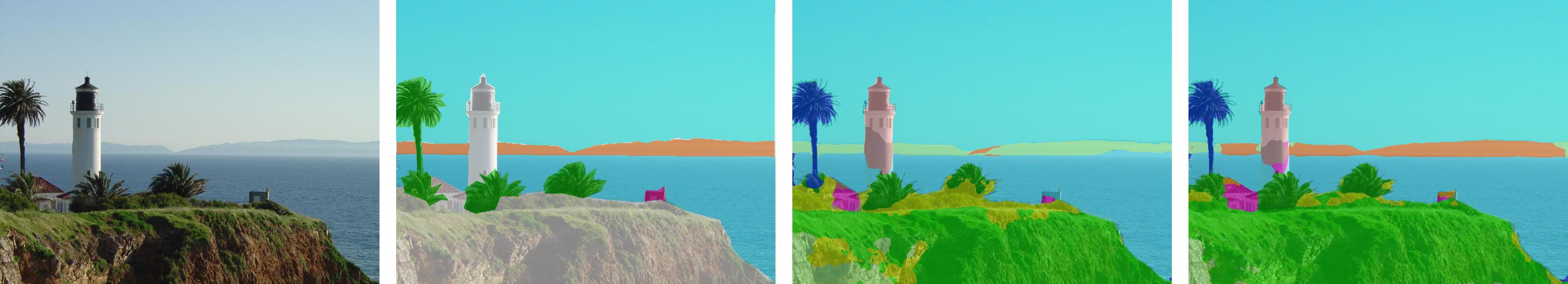}% 

\label{fig:seg}}
\caption{Visualization Results for semantic segmentation on ADE20K. The first column shows the input images. From left to right: Ground Truth, Swin-based, and Iwin-based UperNet.}
\end{figure}

\subsection{Video Recognition on Kinetics-400}

\begin{figure*}[ht]
\centering
\subfloat[]{\includegraphics[width=3.5 in]{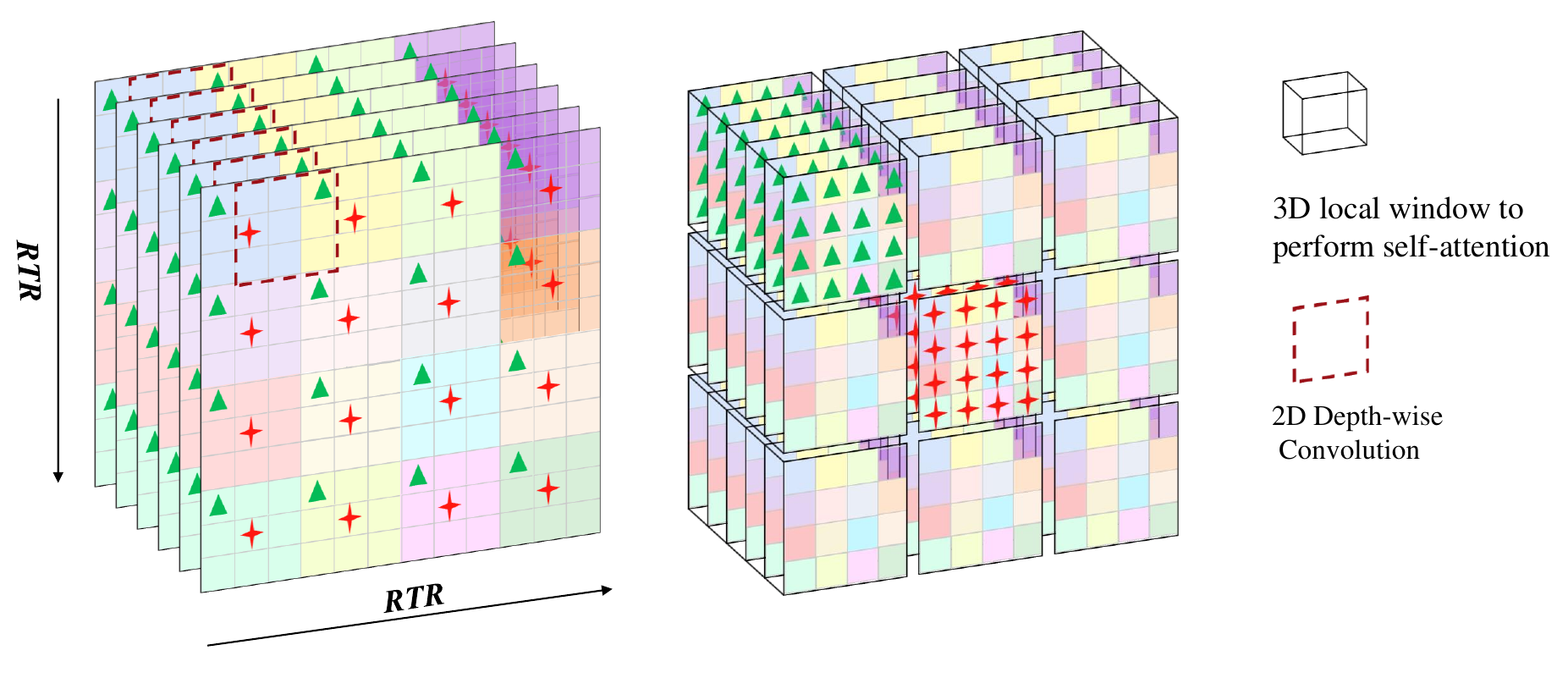}}
\hfil
\subfloat[]{\includegraphics[width=3.5 in]{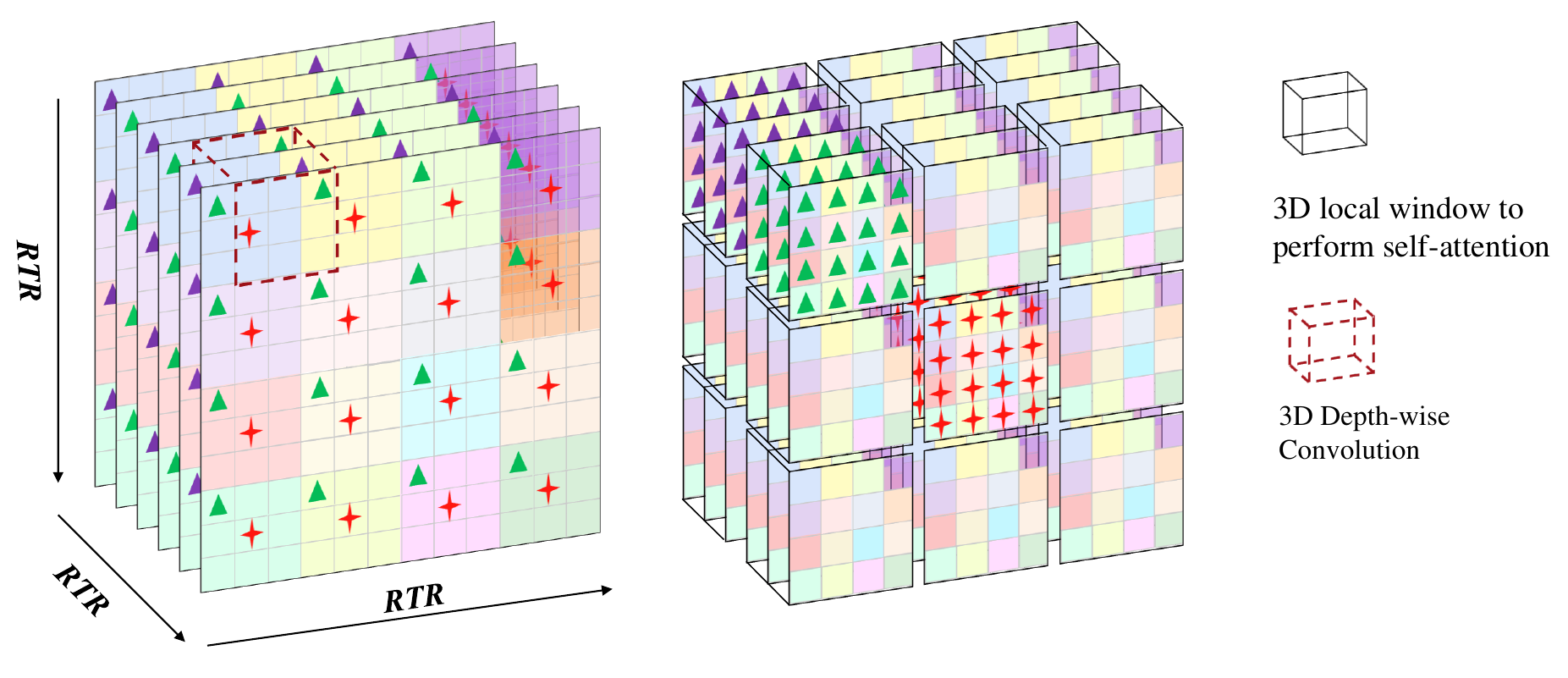}}
\caption{Illustration of Iwin 3D Attention. In (a), 2D depthwise convolution is performed on each frame, and window attention is performed in 3D. Here, in addition to spatial position encoding and collaboration with window attention, the convolution plays a third role: telling the model which tokens are in the same frame, similar to frame attention introduced by VGGT~\cite{wang2025vggt}, thus avoiding temporal position encoding, even though there is no RTR in the temporal dimension. In (b), we additionally perform the RTR operation in the time dimension and cooperate with a 3D depthwise convolution, with its kernel size in the time dimension spanning frames. (a) can be regarded as a special case of (b). This effectively approximates 3D full attention, allowing connections between any tokens within a video. No position coding is required for the time dimension for both methods.}
\label{fig:iwin3d}
\end{figure*}

\paragraph{\textbf{Settings}} Following Video Swin~\cite{liu2021video}, we build Video Iwin for action recognition with MMaction2~\cite{2020mmaction2} on Kinetics-400~\cite{kay2017kinetics}. The 2D ImageNet-pretrained backbone is extended to 3D by adjusting window and kernel sizes only. As in Figure~\ref{fig:iwin3d}, 3D interleaved window attention collects tokens across frames, while depthwise convolution operates per frame (or in 3D) to encode frame membership without explicit temporal position encoding.

\paragraph{\textbf{Results}} According to Table~\ref{tab:k400}, the Iwin model outperforms the Swin in performance and efficiency. At a comparable model scale, Iwin-T achieves a Top-1 accuracy of 79.1\% and a Top-5 accuracy of 93.8\%, slightly surpassing Swin-T’s 78.8\% and 93.6\%. More importantly, Iwin-T’s computational cost is significantly lower, requiring only 74 GFLOPs compared to Swin-T’s 88 GFLOPs, marking a \textbf{15.9}\% reduction. This demonstrates that Iwin-T not only exceeds Swin-T’s performance while offering greater computational efficiency. For the larger-scale Iwin-S, although its Top-1 and Top-5 accuracies (80.0\% and 94.1\%, respectively) are lower than those of Swin-S (80.6\% and 94.5\%), its computational cost (140 GFLOPs) remains considerably lower than Swin-S’s (166 GFLOPs), reflecting a reduction of approximately \textbf{15.7}\%. This indicates that Iwin significantly reduces computational cost while maintaining competitive performance.

\begin{figure}[ht]
\centering
\includegraphics[width=3.5 in]{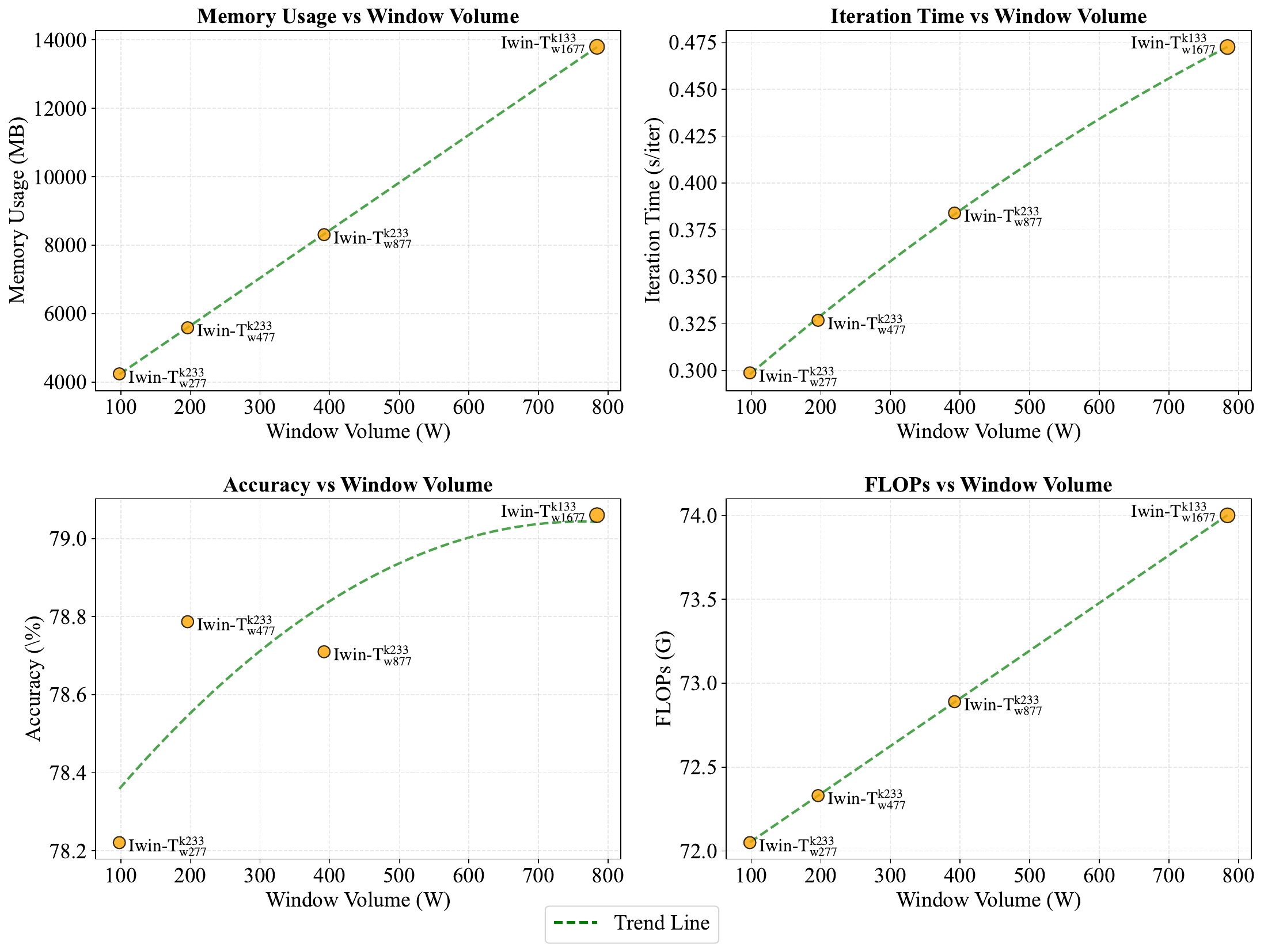}
\caption{Impact of Spatiotemporal Window Volume on Computational Resources and Performance. Larger window volume results in greater computational cost but higher accuracy.}
\label{fig:k400}
\end{figure}

Additionally, we explore the impact of window size on computational resources and performance. The evaluation~\ref{fig:k400} demonstrates a distinct trade-off between computational efficiency and predictive performance across different configurations of the $\mathrm{Iwin\text{-}T}$ architecture, governed by the variations in attention window volume ($W$, defined by window size $w$) and convolution kernel dimensions ($k$). As the spatiotemporal window volume expands from the compact $\mathrm{Iwin\text{-}T}_{\mathrm{w277}}^{\mathrm{k233}}$ ($W=98$) to the extensive $\mathrm{Iwin\text{-}T}_{\mathrm{w1677}}^{\mathrm{k133}}$ ($W=784$), we observe a monotonic increase in computational overhead, where memory usage and iteration latency escalate by approximately $3.2\times$ and $1.6\times$, respectively. While the enlargement of the temporal window in the $\mathrm{w1677}$ variant ($16\times7\times7$) effectively captures longer-range temporal dependencies to achieve the peak accuracy of $79.06\%$, this marginal accuracy gain comes at a disproportionate cost in FLOPs and structural complexity compared to the smaller $\mathrm{w277}$ variant ($78.22\%$). This analysis indicates that while the $\mathrm{Iwin\text{-}T}_{\mathrm{w1677}}^{\mathrm{k133}}$ configuration maximizes precision by leveraging an extensive receptive field, the models with moderate window sizes offer a more balanced efficiency-accuracy profile, minimizing resource consumption while maintaining competitive classification capability.

\begin{table}[t]
\caption{Comparison on Kinetics-400. “Views” indicates \# temporal clip $\times$ \# spatial crop. The magnitudes are Giga ($10^{9}$) and Mega ($10^{6}$) for FLOPs and Param respectively. Iwin-T$^{k233}_{w477}$ means $2\times3\times3$ convolution kernel size and $4\times7\times7$ window size. }
\centering
\resizebox{0.99\columnwidth}{!}{
  \begin{tabular}{l|c|cc|c|cc}
  Method & Pretrain & Top-1 & Top-5 & Views & FLOPs & Param  \\
  \hline
  R(2+1)D~\cite{tran2018closer} & - & 72.0 & 90.0 & 10 × 1 & 75 & 61.8 \\
  I3D~\cite{carreira2017quo} & ImageNet-1K & 72.1 & 90.3 & - & 108 & 25.0 \\
  NL I3D-101~\cite{wang2018non} & ImageNet-1K & 77.7 & 93.3 & 10 × 3 & 359 & 61.8\\
  SlowFast R101+NL~\cite{feichtenhofer2019slowfast} & - & 79.8 & 93.9 & 10 × 3 & 234 & 59.9 \\
  X3D-XXL~\cite{feichtenhofer2020x3d} & - & 80.4 & 94.6 & 10 × 3 & 144 &20.3 \\
  \hline
  MViT-B, 32×3~\cite{fan2021multiscale} & - & 80.2 & 94.4 & 1 × 5 & 170 & 36.6 \\
  MViT-B, 64×3~\cite{fan2021multiscale} & - & 81.2 & 95.1 & 3 × 3 & 455 & 36.6\\
  % TimeSformer-L~\cite{timesformer2021} & ImageNet-21K & 80.7 & 94.7 & 1 × 3 & 2380 & 121.4\\
  % ViT-B-VTN~\cite{neimark2021VTN} & ImageNet-21K & 78.6 & 93.7 & 1 × 1 &4218 & 11.04\\
  ViViT-L/16x2~\cite{arnab2021vivit} & ImageNet-21K & 80.6 & 94.7 & 4 × 3 &1446 & 310.8\\
  ViViT-L/16x2 320~\cite{arnab2021vivit} & ImageNet-21K & 81.3 & 94.7 & 4 × 3 & 3992 & 310.8\\
  \hline
  Swin-T~\cite{liu2021swin} & ImageNet-1K & 78.8 & 93.6 & 4 × 3 & 88 & 28.2 \\
  Swin-S~\cite{liu2021swin} & ImageNet-1K & 80.6 & 94.5 & 4 × 3 & 166 & 49.8 \\
  % \hline
 
  Iwin-T$^{k233}_{w277}$(ours) & ImageNet-1K & \textbf{78.2} & \textbf{93.8} & 4 × 3 & \textbf{72.05} & 29.8 \\
  Iwin-T$^{k233}_{w477}$(ours) & ImageNet-1K & \textbf{78.8} & \textbf{93.8} & 4 × 3 & \textbf{72.33} & 29.8 \\
   Iwin-T$^{k233}_{w877}$(ours) & ImageNet-1K & \textbf{78.7} & \textbf{93.8} & 4 × 3 & \textbf{72.89} & 29.8 \\
    Iwin-T$^{k133}_{w1677}$(ours) & ImageNet-1K & \textbf{79.1} & \textbf{93.8} & 4 × 3 & \textbf{74} & 29.8 \\
%   \hline
  Iwin-S$^{k133}_{w1677}$(ours) & ImageNet-1K & 80.0 & 94.1 & 4 × 3 & \textbf{140} & 51.1 \\ 
  \end{tabular}}
\label{tab:k400}
\end{table}

\subsection{Image Generation on ImageNet-1K}
\label{sec:imgen}

\paragraph{\textbf{Settings}} Following LightningDiT~\cite{yao2025vavae}, we build FlashDiT to validate the Iwin module for class-conditional image generation on ImageNet. Standard self-attention is replaced by interleaved window attention plus depthwise convolution; position encodings are removed. Following the design guideline $KM \geq \max(H,W)$ on a flat $16{\times}16$ latent ($256{\times}256$ input with $16{\times}$ downsampling), we set $k{=}3$ and evaluate $M{\in}\{4,8\}$ ($3{\times}4{=}12{\approx}16$; $3{\times}8{=}24{>}16$). The attention and convolution branches receive \emph{separate} adaptive layer-norm (adaLN) time-step modulations (scale and shift), respectively, enabling independent conditioning of each pathway.

\begin{figure}[ht]
\centering
\includegraphics[width=3.5 in]{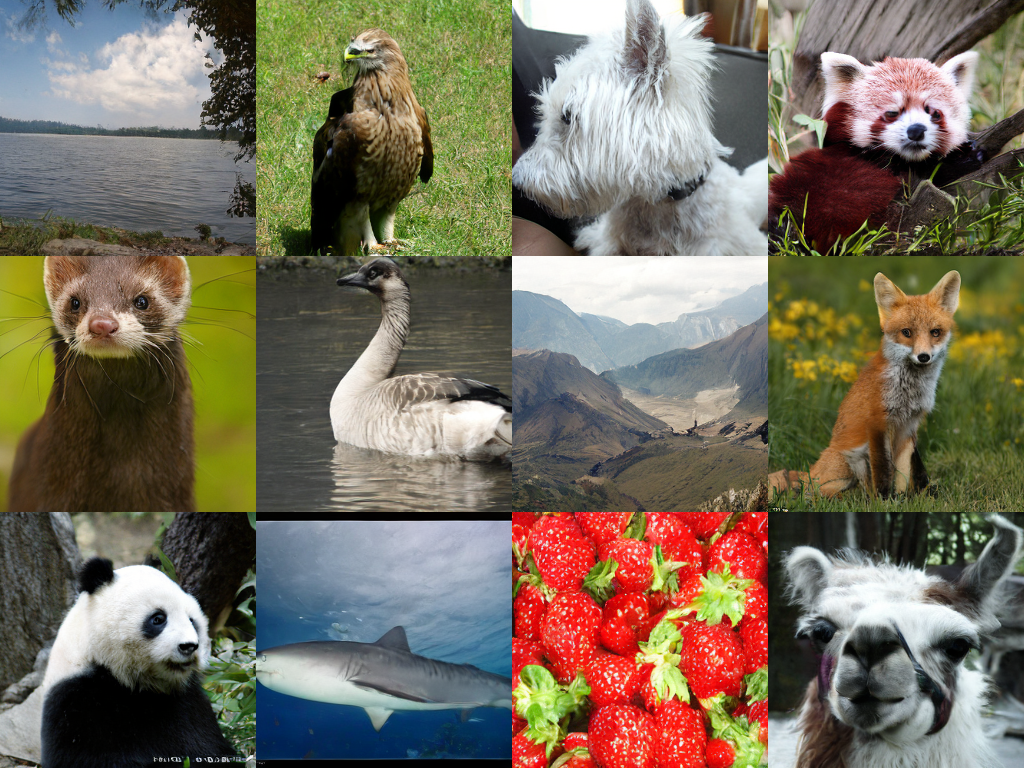}
\caption{\textbf{Visualization Results.} Some sampled images at 256 × 256 resolution generated by the trained Flashdit(win8) model.
}
\label{fig:samples}
\end{figure}

\paragraph{\textbf{Results}} Table~\ref{tab:imgen} compares FlashDiT with prior generators. Under the same 675M parameter budget, FlashDiT(win8) reaches w/o CFG gFID 5.37 and w/ CFG 2.30 after only 56 training epochs---far fewer than DiT/SiT (1400), MaskDiT (1600), or MAR (800)---yet remains competitive: w/o CFG it clearly beats DiT (9.62) and SiT (8.61) and is on par with MaskDiT (5.69); w/ CFG it matches DiT (2.27) and MaskDiT (2.28). Enlarging the window from win4 to win8 further improves quality (w/o CFG gFID 7.08$\rightarrow$5.37, IS 120.8$\rightarrow$131.4; w/ CFG gFID 2.90$\rightarrow$2.30, IS 229.2$\rightarrow$244.1), indicating that it is best to satisfy $KM \geq \max(H, W)$. We did not pursue longer schedules due to compute limits and extended training should narrow the gap to fully trained diffusion baselines. Samples from FlashDiT(win8) are in Figure~\ref{fig:samples}.

\begin{table*}[t]
\centering
\caption{Comparison with state-of-the-art models on ImageNet for image generation.}
% \resizebox{0.99\columnwidth}{!}{
\begin{tabular}{l|cc|ccccc|ccccc}
\toprule
\multirow{2}{*}{\textbf{Method}} & \multirow{2}{*}{\textbf{Epoches}} & \multirow{2}{*}{\textbf{Params}} & \multicolumn{5}{c|}{\textbf{256$\times$256 w/o CFG}} & \multicolumn{5}{c}{\textbf{256$\times$256 w/ CFG}} \\
\cmidrule(lr){4-8} \cmidrule(lr){9-13}
& & & \textbf{gFID} $\downarrow$ & \textbf{sFID} $\downarrow$ & \textbf{IS} $\uparrow$ & \textbf{Pre.} $\uparrow$ & \textbf{Rec.} $\uparrow$ & \textbf{gFID} $\downarrow$ & \textbf{sFID} $\downarrow$ & \textbf{IS} $\uparrow$ & \textbf{Pre.} $\uparrow$ & \textbf{Rec.} $\uparrow$ \\
\midrule
\multicolumn{13}{c}{\textit{\textbf{AutoRegressive (AR)}}} \\
\midrule
MaskGIT~\cite{maskgit} & 555 & 227M & 6.18 & - & 182.1 & 0.80 & 0.51 & - & - & - & - & - \\
LlamaGen~\cite{llamagen} & 300 & 3.1B & 9.38 & 8.24 & 112.9 & 0.69 & 0.67 & 2.18 & 5.97 & 263.3 & 0.81 & 0.58 \\
VAR~\cite{var} & 350 & 2.0B & - & - & - & - & - & 1.80 & - & 365.4 & 0.83 & 0.57 \\
MagViT-v2~\cite{magvitv2} & 1080 & 307M & 3.65 & - & 200.5 & - & - & 1.78 & - & 319.4 & - & - \\
MAR~\cite{mar} & 800 & 945M & 2.35 & - & 227.8 & 0.79 & 0.62 & 1.55 & - & 303.7 & 0.81 & 0.62 \\
\midrule
\multicolumn{13}{c}{\textit{\textbf{Latent Diffusion Models}}} \\
\midrule
MaskDiT~\cite{maskdit} & 1600 & 675M & 5.69 & 10.34 & 177.9 & 0.74 & 0.60 & 2.28 & 5.67 & 276.6 & 0.80 & 0.61 \\
DiT~\cite{peebles2023scalable} & 1400 & 675M & 9.62 & 6.85 & 121.5 & 0.67 & 0.67 & 2.27 & 4.60 & 278.2 & 0.83 & 0.57 \\
SiT~\cite{sit} & 1400 & 675M & 8.61 & 6.32 & 131.7 & 0.68 & 0.67 & 2.06 & 4.50 & 270.3 & 0.82 & 0.59 \\
MDTv2~\cite{mdtv2} & 1080 & 675M & - & - & - & - & - & 1.58 & 4.52 & 314.7 & 0.79 & 0.65 \\
REPA~\cite{repa} & 800 & 675M & 5.90 & - & - & - & - & 1.42 & 4.70 & 305.7 & 0.80 & 0.65 \\
LightningDiT~\cite{yao2025vavae} & 64 & 675M & 5.14 & 4.22 & 130.2 & 0.76 & 0.62 & 2.11 & 4.16 & 252.3 & 0.81 & 0.58 \\
\midrule
% FlashDiT(win4, Ours) & 56 & 675M & 7.88 & 5.91 & 116.2 & 0.73 & 0.60 & 3.08 & 6.00 & 223.2 & 0.78 & 0.57 \\
FlashDiT(win4, Ours) & 56 & 675M & 7.08 & 5.68 & 120.8 & 0.74 & 0.61 & 2.90 & 5.71 & 229.2 & 0.78 & 0.57 \\
FlashDiT(win8, Ours) & 56 & 675M & 5.37 & 4.62 & 131.4 & 0.76 & 0.61 & 2.30 & 4.53 & 244.1 & 0.81 & 0.58 \\
\bottomrule
\end{tabular}
% }
\label{tab:imgen}
\end{table*}

\begin{figure}[ht]
\centering
\includegraphics[width=3.5 in]{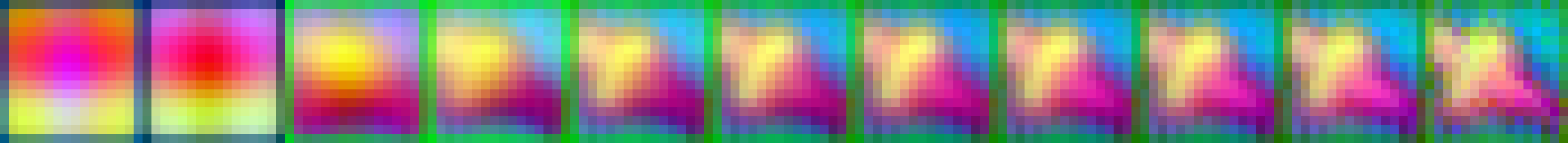}
\includegraphics[width=3.5 in]{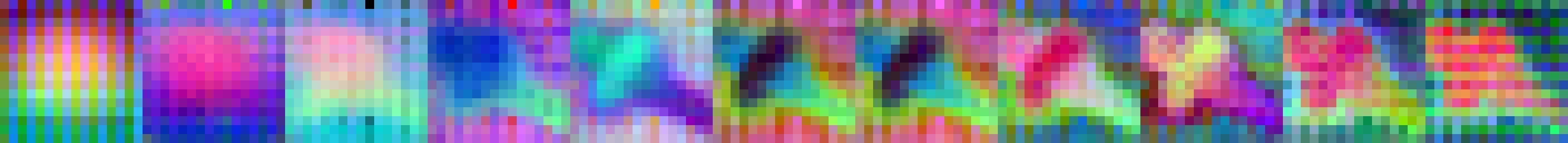}
\includegraphics[width=3.5 in]{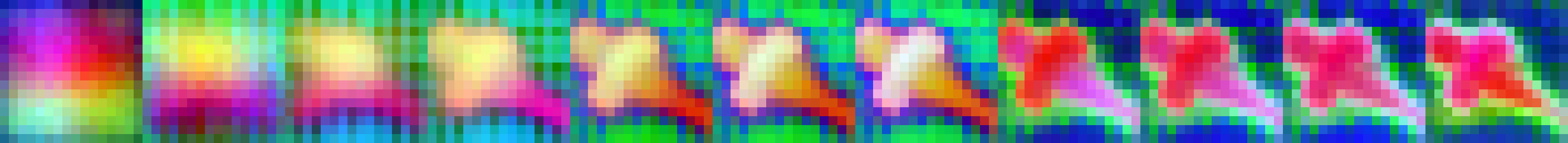}
\caption{PCA visualization of the latent sampled during the diffusion process for category 0816. Row 1 shows the convolutional component capturing high‑frequency textures, row 2 the attention component encoding contours, and row 3 the fused output yields a complete latent.
}
\label{fig:pca}
\end{figure}

\subsection{Ablation Study}

\begin{table}[ht]
  \centering
  \caption{Ablation studies on various architectural components. FLOPs are Giga ($10^9$) and Param are Mega ($10^6$).}
  \resizebox{\columnwidth}{!}{
  \begin{tabular}{l|ccc|c}
  \toprule
  \textbf{Setting} & \textbf{Param(M)} & \textbf{FLOPs(G)} & \textbf{Throughput(img/s)} & \textbf{Top-1 Acc(\%)} \\
  \midrule
  \multicolumn{5}{c}{\textbf{Ablation on Attention and Convolution Combination}} \\
  DwConv           & 21.60        & 3.20         & 861         & 79.4          \\
  W-MSA            & 30.20        & 4.71         & 758         & 80.2          \\
  IW-MSA           & 30.20        & 4.71         & 756         & 80.4          \\
  DwConv + W-MSA   & 30.23        & 4.72         & 737         & 81.8          \\
  DwConv + IW-MSA  & 30.23        & 4.72         & 736         & \textbf{82.0} \\
  \midrule
  \multicolumn{5}{c}{\textbf{Ablation on Downsampling Methods}} \\
  % \midrule
  DWConv           & 27.14        & 4.57         & 746         & 81.9          \\
  Avg Pooling      & 27.13        & 4.51         & \textbf{762} & 81.8          \\
  Patch Merging    & 28.29        & 4.51         & 741         & 81.8          \\
  Std Conv         & 30.23        & 4.72         & 736         & \textbf{82.0} \\
  \midrule
  \multicolumn{5}{c}{\textbf{Ablation on Kernel Size for Depthwise Convolution}} \\
   % \midrule
  \{7, 5, 3, None\} & 30.24        & 4.75         & 714         & 81.0          \\
  \{7, 7, 7, None\} & 30.34        & 4.78         & 729         & 82.1          \\
  \{5, 5, 5, None\} & 30.27        & 4.75         & 731         & \textbf{82.2} \\
  \{3, 3, 3, None\} & 30.23        & 4.72         & \textbf{736} & 82.0          \\
  \midrule
  \multicolumn{5}{c}{\textbf{Ablation on Block Number Configuration}} \\
   % \midrule
  \{4, 3, 2, 2\}   & 23.79        & 4.43         & 473         & 80.5          \\
  \{3, 3, 3, 3\}   & 32.56        & 4.80         & 588         & 81.8          \\
  \{2, 2, 6, 2\}   & 30.23        & 4.72         & \textbf{736} & \textbf{82.0} \\
  \midrule
  \multicolumn{5}{c}{\textbf{Ablation on Position Embedding}} \\
   % \midrule
  abs. pos.  & 30.53        & 4.72         & 735         & 82.1          \\
  rel. pos.  & 30.25        & 4.72         & 724         & \textbf{82.4}          \\
  no pos.    & 30.23        & 4.72         & \textbf{736} & 82.0 \\
  rel. pos. (Iwin-S)  & 51.60        & 8.98         & 403         & 83.3          \\
  no pos. (Iwin-S)   & 51.60         & 8.98         & \textbf{410} & \textbf{83.4} \\
  \bottomrule
  \end{tabular}}
  \label{tab:ablations}
\end{table}

Our extensive ablation studies, summarized in Table~\ref{tab:ablations}, are all conducted on Iwin-T and ImageNet by default.

\paragraph{\textbf{Attention and Convolution Combination}} 
We investigated the effects of integrating depthwise separable convolutions (DWConv) and different attention mechanisms. As demonstrated, the collaboration of DWConv and IW-MSA (Interleaved Window Multi-head Self-Attention) yields the best performance, achieving a Top-1 accuracy of 82.0\%. This verified the superiority of our proposed method.

\paragraph{\textbf{Downsampling Methods}}
We evaluated various downsampling methods between network stages. The results show that using Standard Convolution (Std Conv), adopted in our final Iwin design, achieves the highest accuracy of 82.0\%. We struggled between standard convolution and average pooling offering highest throughput, but for higher accuracy, we chose standard convolution.

\paragraph{\textbf{Kernel Size Choice}}
We explored the impact of varying kernel sizes for DWConv across different stages. Our final configuration, which utilizes a fixed kernel size of \{3, 3, 3, None\}, yields a accuracy of 82.0\% with a throughput of 736 img/s. While a \{5, 5, 5, None\} kernel size achieves a slightly higher accuracy of 82.2\%, the smaller \{3, 3, 3, None\} kernels offer a more favorable trade-off between performance and computational efficiency (higher throughput). Additionally, using different kernel sizes at different stages to satisfy $KM \geq \max(H,W)$ did not yield the best results. This aligns with observations in \cite{tan2019efficientnet} that balanced network outperform theoretically optimal but imbalanced configurations.

\paragraph{\textbf{Block Number Configuration}}
We examined different distributions of block numbers across the four stages of the network. Initially, we explored the \{4, 3, 2, 2\} configuration, which aimed to approximate larger kernel sizes (e.g., a $7\times7$ kernel with four $3\times3$ convolutions) by stacking smaller convolutional blocks. However, this setup resulted in the lowest accuracy (80.5\%) and throughput (473 img/s). In contrast, the \{2, 2, 6, 2\} configuration achieved the highest accuracy of 82.0\% with a throughput of 736 img/s. This configuration, which allocates more blocks to the deeper stages, proves beneficial for maximizing performance while maintaining optimal speed.

\paragraph{\textbf{Position Embedding}}
We explored the effect of positional encoding on the models. In the Iwin-T model, relative position embedding achieves the highest accuracy at 82.4\%. However, in deeper models like Iwin-S, the no-position-embedding approach reaches the highest Top-1 accuracy of 83.4\%, surpassing the relative position embedding (83.3\%) and processing more images per second (410 img/s). This indicates that in very deep networks, position embeddings might be unnecessary or even harmful. Additionally, during training, models with absolute or relative position embeddings take more time to learn compared to those without position embedding.

\begin{figure*}[ht]
\centering
\includegraphics[width=\textwidth]{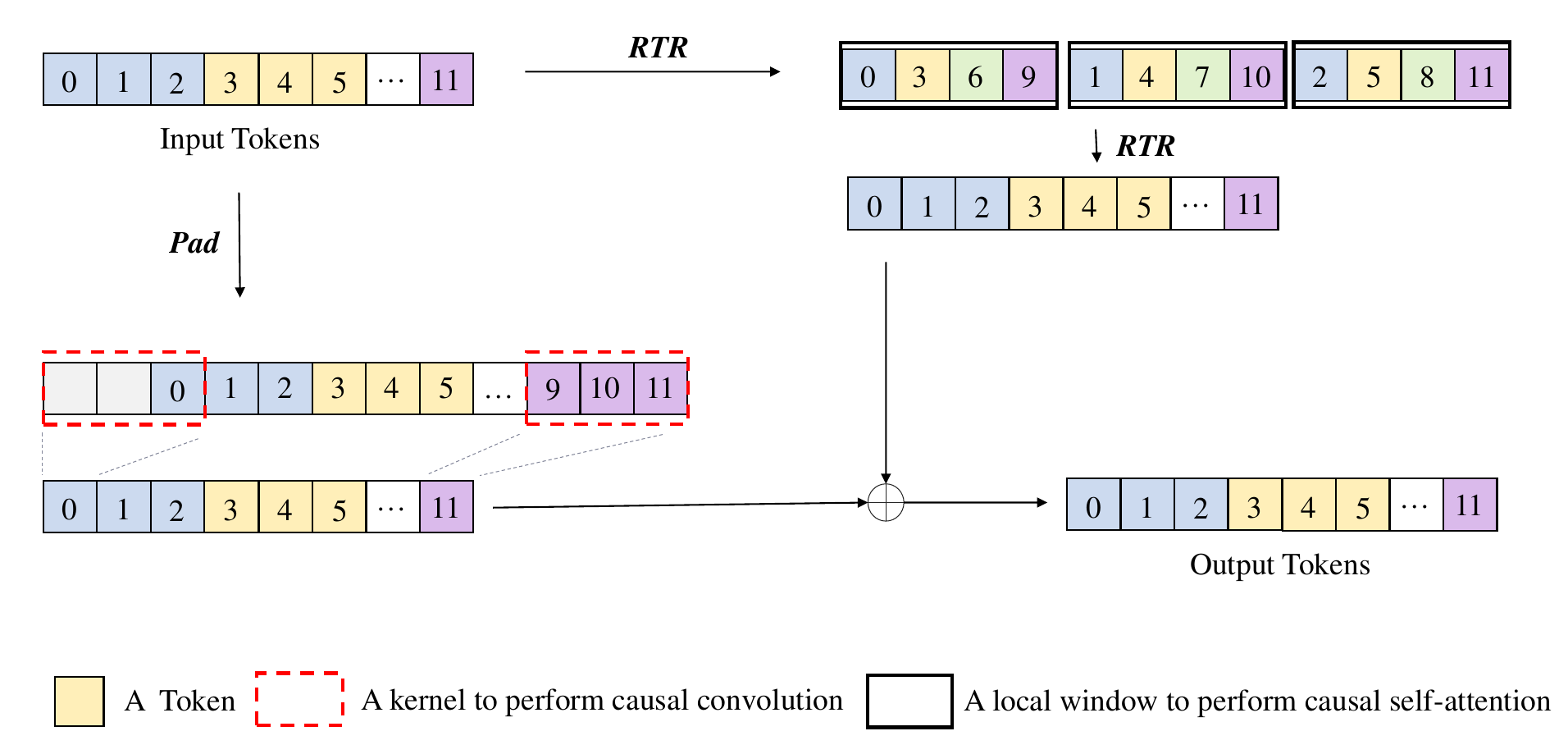}
\caption{The illustration of Iwin 1D Attention, shows how to the apply Iwin Transformer concepts into large language models (LLMs). The red boxes represent causal depthwise separable convolution, while the black boxes denote causal attention within a window. Neither process leaks future information. The final output, derived from the combination of these two causal operations, adheres to the principle of causality. This approach might be also helpful in solving the problem of high complexity with long sequences in LLMs.}
\label{fig:nlp}
\end{figure*}

\section{Discussion}
\label{sec:discussion}
We believe that some fields of work can benefit and be inspired by Iwin Transformer.

\subsection{Migration to Large Language Models}

The Iwin Transformer's position-embedding-free design principle offers promising opportunities for application in Large Language Models (LLMs). Currently, LLMs heavily depend on position embeddings to preserve sequence order information. By integrating interleaved window attention with depthwise separable convolution, it may be possible to achieve more natural length generalization. This approach relies on structural rather than parametric position information, facilitating easier length extrapolation. As shown in Figure~\ref{fig:nlp}, the computation is divided into two components: 1D causal depthwise convolution and 1D interleaved window causal attention. Both ensure tokens relate only to preceding tokens, giving Iwin 1D Attention causality.

\subsection{Application to Generation Models}
Generation Models, such as diffusion models, present another domain where Iwin's design could offer substantial benefits. Iwin's absence of position embedding facilitates seamless adaptation to varying resolutions without parameter interpolation, which is valuable for progressive or multi-scale generation. The faster convergence observed with FlashDiT suggests that the convolutional inductive bias can shorten diffusion training. Iwin 3D Attention offers a third option for video generation alongside expensive 3D full attention and serial spatiotemporal attention: a single 3D window attention plus depthwise convolution relates all tokens jointly, potentially avoiding the distribution mismatch that can arise when spatial and temporal attention are applied sequentially.

\subsection{Multi-View Video Modeling}
Autonomous driving and robotics with multi-camera have input tensors of shape $(B, N, T, H, W, C)$, where $N$ is the number of synchronized views. We can apply 3D depthwise convolution on $(B{\cdot}N, T, H, W, C)$ to capture local associations across views, time, and space. RTR is then performed on $(T,H,W)$, and interleaved windows are partitioned into $(B{\cdot}W_t W_h W_w,\; N T_w H_w W_w, C)$ groups for IW-MSA. Convolution encodes view, frame, and spatial neighborhoods, so no learnable view-index or temporal position embeddings are required, as has already been verified in video tasks. So we can build a positional-embedding-free 4D model. With window sizes $W_t{=}W_h{=}W_w{=}2$, the effective token count per attention window is reduced by $1/8$; with $W{=}8$ along each dimension, the reduction reaches $1/512$, offering substantial savings for long multi-view sequences.

\subsection{Progressive 2D-to-3D-to-4D Transfer}

A further benefit of Iwin's position-embedding-free design is \emph{progressive dimensional transfer}: higher-dimensional models can inherit weights from lower-dimensional pre-training without redefining or interpolating position parameters. Our Kinetics-400 experiments already validate the 2D$\rightarrow$3D transfer; a Video Iwin model is initialized from an ImageNet-pretrained 2D Iwin by adjusting window and kernel sizes only. The same logic extends to 4D tensors $(B,N,T,H,W,C)$ : a multi-view model can be warm-started from a 3D video checkpoint and, in turn, from a 2D image checkpoint, substantially reducing training time by injecting strong spatio-temporal priors at each stage. Iwin is naturally suited to this curriculum because every dimensional lift reuses the \emph{same} IW-MSA+DWConv block; only the RTR axes and window shapes change.

\section{Conclusion}
\label{sec:conclusion}

In this paper, we introduced Iwin Transformer, a position-embedding-free vision transformer that couples interleaved window attention with depthwise convolution playing multiple roles. This design simultaneously reduces the quadratic complexity of attention and unlocks two forms of scalability: across resolutions, it allows direct fine-tuning by simply adjusting window/kernel sizes without positional interpolation needed; across dimensions, the same block lifts from 2D to 3D video (validated on Kinetics‑400) and holds the potential for 4D World Models. We hope this unified design helps large-scale models and high‑resolution tasks tame the compute growth along spatial, temporal, and multi-view axes, making scalable vision transformers more practical for the next generation of generative and embodied applications.

\bibliographystyle{IEEEtran}
\bibliography{ref}

\vfill

\end{document}